\documentclass{article}

\usepackage[letterpaper,top=2cm,bottom=2cm,left=3cm,right=3cm,marginparwidth=1.75cm]{geometry}

\usepackage[colorlinks=true, allcolors=blue]{hyperref}

\usepackage{tabularx}
\usepackage{booktabs}
\usepackage{graphicx}
\usepackage{bm}
\usepackage{enumitem}
\usepackage{algorithm}
\usepackage{algpseudocode}
\usepackage{subfigure}
\usepackage{subcaption}

\usepackage{amsmath}
\usepackage{amssymb}
\usepackage{mathtools}
\usepackage{amsthm}
\usepackage{wrapfig}
\usepackage{mdframed}
\usepackage[numbers]{natbib}

\newtheorem{definition}{Definition}[section]
\newtheorem{lemma}{Lemma}[section]

\newcommand{\mat}[1]{\mathbf{#1}}

\usepackage{tikz}
\usepackage{xcolor}

\title{Poor Man's Training on MCUs: A Memory-Efficient Quantized Back-Propagation-Free Approach}
\author{
Yequan Zhao\thanks{\small{Department of Electrical and Computer Engineering, UC Santa Barbara, Santa Barbara, California, USA}}
\quad
Hai Li\thanks{\small{Intel Corporation, Hillsboro, Oregon, USA} }
\quad
Ian Young\footnotemark[2]
\quad
Zheng Zhang\footnotemark[1]
}
\begin{document}
\maketitle

\begin{abstract}
Back propagation (BP) is the default solution for gradient computation in neural network training. However, implementing BP-based training on various edge devices such as FPGA, microcontrollers (MCUs), and analog computing platforms face multiple major challenges, such as the lack of hardware resources, long time-to-market, and dramatic errors in a low-precision setting. This paper presents a simple BP-free training scheme on an MCU, which makes edge training hardware design as easy as inference hardware design. We adopt a quantized zeroth-order method to estimate the gradients of quantized model parameters, which can overcome the error of a straight-through estimator in a low-precision BP scheme. We further employ a few dimension reduction methods (e.g., node perturbation, sparse training) to improve the convergence of zeroth-order training. Experiment results show that our BP-free training achieves comparable performance as BP-based training on adapting a pre-trained image classifier to various corrupted data on resource-constrained edge devices (\textit{e.g.,} an MCU with 1024-KB SRAM for dense full-model training, or an MCU with 256-KB SRAM for sparse training). This method is most suitable for application scenarios where memory cost and time-to-market are the major concerns, but longer latency can be tolerated. 
\end{abstract}

\section{Introduction}
On-device training, that is training deep neural networks (DNN) on edge devices, enables a DNN model pre-trained on \textit{cloud} to improve itself on newly observed data and adapt to cross-domain or out-of-domain distribution shifts after edge deployment. It also allows the model to adapt to user personalization locally, which protects user privacy over sensitive data (e.g., healthcare and financial data). As physic-informed machine learning has been increasingly used for safety-critical decision-making in autonomous systems, there has been also growing interest in on-device fine-tuning or end-to-end training. In federated learning, a machine learning model also needs to be trained periodically on each local edge node, then updated on a global centralized server. 

Backward propagation (BP)~\cite{lecun1988theoretical} is used in almost all neural network training frameworks for gradient computation. BP is actually a reverse-mode automatic differentiation (AD)~\cite{baydin2018automatic,margossian2019review} approach implemented based on the information of a computational graph. While a forward-mode AD is suitable for computing the gradient of a single-input multiple-out function, BP is more suitable for a multiple-input (i.e., many network parameters) and single-output (i.e., training loss) function. With sophisticated AD packages, operating systems, and compilers, BP can be called with just one command (e.g., {\texttt loss.backward()} in PyTorch) on a CPU- or GPU-based desktop or cloud computing platform. This has greatly simplified the development and deployment of modern neural network models. 

However, training a neural network on resource-constrained edge hardware [e.g., a microcontroller unit (MCU), FPGA or photonic platform] is completely different from the training task on a desktop or cloud platform, due to the limited hardware resources and software support. Specifically, implementing a standard BP-based training framework on edge devices are often prevented by three major challenges:

\begin{itemize}[leftmargin=*]
\item {\bf Memory Challenge.} Edge devices like MCU have a very limited run-time memory (\textit{e.g.,} STM32F746 with only 256-KB user SRAM, or STM32H7B3 with 1024-KB user SRAM). This budget is often below the memory requirement of storing all network parameters, making full-model BP-based training impossible for most realistic cases. 
 By choosing tailored network models (\textit{e.g.,} MCUNet \cite{lin2020mcunet}), using real-quantized graphs and a co-designed lightweight back-end (\textit{e.g.,} the TinyEngine \cite{lin2020mcunet, lin2022device}), one may perform
edge inference with a low memory cost (e.g., 96 KB for the MCUNet-in1 model~\cite{lin2020mcunet}). However, the memory cost of a full-model BP-based training (e.g., 7.4 MB for MCUNet-in1) is far beyond the memory capacity.
Existing training methods on MCU update only a small subset of model parameters (e.g., only the last layer~\cite{mudrakarta2018k, ren2021tinyol}, bias vectors~\cite{cai2020tinytl} to reduce the memory cost, yet leads to significant (e.g., >10\%) accuracy drop. Sparse update~\cite{lin2020mcunet, kwon2023tinytrain}) could narrow this gap, yet requires computation-intensive searches and compilation-level optimization on \textit{cloud}.

\item {\bf Precision Challenge.} Low-precision quantized computation is often utilized on digital edge hardware (e.g., MCU and FPGA) to reduce latency, memory cost, and energy consumption. However, low-precision operations pose great challenges for BP-based training. BP was originally designed for the gradient computation of smooth functions. Thus, error-prone approximation techniques such as straight-through estimators \cite{bengio2013estimating} are required to handle non-differentiable functions in quantized neural network training. The errors introduced by these approximation techniques increase as hardware precision reduces. They also propagate and accumulate through different layers, leading to dramatic accuracy drop, unstable training behaviors, or even divergence \cite{zhu2020towards, chen2020statistical}.


\item {\bf Time-to-Market Challenge.}  While BP can be done on CPU or GPU with just one line of code (e.g., \texttt{loss.backward()} in PyTorch), implementing it on edge devices can be very challenging. Due to the lack of automatic differentiation packages~\cite{baydin2018automatic} and sophisticated operating systems on edge platforms, designers often have to implement the math and hardware of gradient computation manually. On some platforms (e.g., integrated photonics), novel devices must be invented and fabricated to perform BP~\cite{pai2023experimentally}. This error-prone process needs numerous debugs and design trade-offs.  As a result, designing edge training hardware is more time-consuming than designing inference hardware. For instance, our own experience shows that an experienced FPGA designer can design a high-quality inference accelerator within one week, yet it takes over one year to implement an error-free training accelerator on FPGA. This long time to market is often unacceptable in the industry due to the fast evolution of AI models. 
\end{itemize}


{\bf Paper Contributions.} The above challenges motivate us to ask the following question:

\begin{mdframed}[userdefinedwidth=5.8inch, align=center]
 Can we make the edge training hardware design as easy and memory-efficient as inference hardware design?
\end{mdframed}
In this paper, we show that the answer is affirmative, with the assumption that memory budget and time to market are given higher priority over runtime latency. Our key idea is to completely bypass the complicated BP implementation by proposing a \textbf{quantized zeroth-order (ZO) method} to train a real-quantized neural network model on MCU. This training method only uses quantized forward evaluations to estimate gradients. As a result, we can use a similar memory cost of inference to achieve full-model training under the tiny memory budget of an MCU. This quantized ZO training framework can be used as a plug-and-play tool added to quantized inference hardware, therefore the design complexity and time to market can be dramatically reduced.  Our specific  contributions are briefly summarized below:
\begin{enumerate}
    \item {\bf ZO Quantized Training for Edge Devices.} We propose a BP-free training framework via quantized zeroth-order optimization to enable full-model and real-quantized training on MCUs under extremely low memory budget (\textit{e.g.,} 256-KB SRAM for sparse training or 1024-KB SRAM for dense training). This framework enjoys low memory cost and easy implementation. Furthermore, it shows better accuracy than quantized BP-based training in low-precision (e.g., INT8) settings since no error-prone straight-through estimator is needed. 
    \item {\bf Convergence Improvement.} ZO training suffers from slow convergence rates as the number of training variables increases. Previous assumption of low intrinsic dimensionality~\cite{malladi2023fine} or  coordinate-wise gradient estimation \cite{chen2023deepzero} does not work in on-device training. To improve the training convergence, we propose a learning-rate scaling method to stabilize each training step. We also employ a few dimension-reduction methods to improve the training convergence: (i) a generic layer-wise gradient estimation strategy that combines weight perturbation and node perturbation for ZO gradient estimation, (ii) a sparse training method with task-adaptive block selection to reduce the number of the trainable parameters.
    \item {\bf MCU Implementation.} We implement the proposed BP-free training framework on an MCU (full-model training on an STM32H7B3 with 1024-KB user SRAM, and sparse training on an STM32F746 with 256-KB user SRAM). A quantized inference engine is easily converted to a training engine, with only an additional control unit, a temporary gradient buffer, and a pseudo-random number generator. To our best knowledge, this is the first framework to enable full-model training under such a tiny memory budget (STM32H7B3 with 1024-KB user SRAM). 
    \item {\bf Experimental Validation.} We conduct extensive experiments on adapting a pre-trained image classification model to unseen image corruptions and fine-grained vision classification datasets. On adapting image corruptions, our BP-free training outperforms current quantized BP-based training with an average 6.4\% test accuracy improvement. Our method can also match the performance of back-propagation training on fine-grained vision classification datasets. 
\end{enumerate}
\begin{figure}[t]
    \centering
    \includegraphics[width=0.9\linewidth]{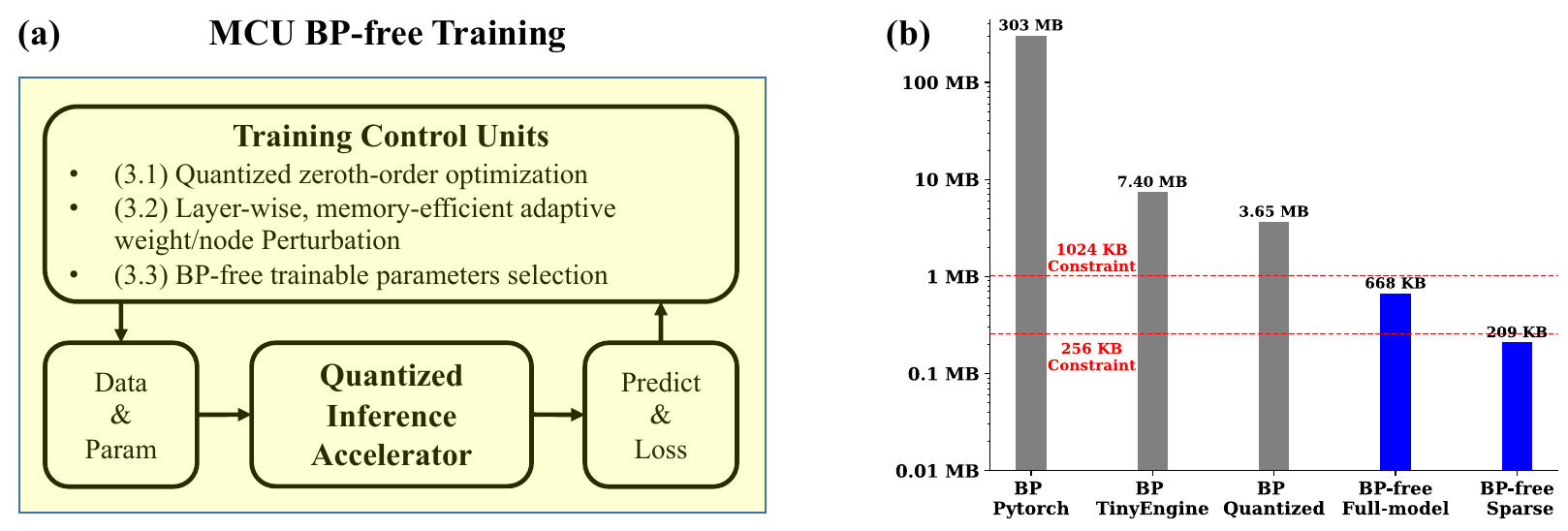}
    \caption{(a): Overview of BP-free training framework. A quantized inference engine is easily converted to a training engine by adding control unit and repeatedly calling the inference accelerator. (b): Training memory comparison of different training methods. The numbers are measured with MCUNet-in1~\cite{lin2020mcunet}, batch size 1, and resolution 128$\times$ 128.
    }
    \label{fig:main_concept}
\end{figure}

Our Key idea is summarized in Fig.~\ref{fig:main_concept} (a). As demonstrated in Fig.~\ref{fig:main_concept} (b), our method  is the only solution to enable full-model training on commodity-level MCU (\textit{e.g.,} STM32H7B3 with 1024-KB user SRAM) without auxiliary memory. 
This memory cost is the minimum to enable full-model training (478 KB model parameters plus 190 KB peak inference memory), $5.46\times$ more memory-efficient than the memory cost of quantized BP, and $>400\times$ more memory efficient than BP-based training in PyTorch which includes back-end memory overhead. 
BP-free sparse training further reduces the memory cost to fit a smaller budget (\textit{e.g.,} 256-KB SRAM).

\section{Preliminaries}
\subsection{Real-quantized Neural Network Model}
Given a full-precision linear layer $\mat{z}=\mat{W}\mat{x}+\mat{b}$, the low-precision (e.g., INT8) quantized counterpart is:
\begin{equation}
    \bar{\mat{z}} = \text{clip}(\lceil (s_{\mat{W}}s_{\mat{x}}\bar{\mat{W}}\bar{\mat{x}} + \bar{\mat{b}}) / s_{\mat{z}} \rfloor, -Q_N, Q_P)
\end{equation}
Here $\bar{\cdot}$ denotes the quantized variables, $s$ is a floating-point scaling factor, $\text{clip}(v,r_1,r_2)$ returns $r_1$ when $v\leq r_1$ (or $r_2$ when $v\geq r_2$), and $\lceil v \rfloor$ rounds $v$ to the nearest integer. Given a $b$-bit integer format, $Q_N=2^{b-1}$ and $ Q_P=2^{b-1}-1$  for signed quantization; $Q_N=0$ and $ Q_P=2^{b}-1$ for unsigned quantization. We call it a \textit{real-quantized} graph \cite{lin2022device} since all variables are in low-precision formats, and all matrix multiplications are computed with fixed-point arithmetic (the scaling factor $s$ could further be quantized to achieve only fixed-point multiplication \cite{jin2022f8net}). On resource-constrained edge devices, \textit{real-quantized} graphs are usually leveraged to achieve memory and computation efficiency. 

In this paper, we focus on the training of a \textit{real-quantized} neural network on MCU. We remark that training on a real-quantized graph fundamentally differs from quantization-aware training~\cite{jacob2018quantization} and fully-quantized training~\cite{chen2020statistical}: the latter two maintain a full-precision copy of model parameters and only quantize model parameters and activations to leverage fixed-point computation. The comparison can be found in Table \ref{tab:quantized training}. 


\begin{table}[htbp]
  \centering
  \caption{Comparison of different quantized training paradigms. 
  }
    \begin{tabular}{cccc}
    \toprule
    \toprule
          & Model & Forward & Backward \\
    \midrule
    Quantization-aware Training (QAT) \cite{jacob2018quantization} & FP  & INT   & FP \\
    Fully-quantized Training (FQT) \cite{chen2020statistical} & FP  & INT   & INT \\
    \textbf{Real-quantized Training (RQT)} \cite{lin2022device} & INT   & INT   & INT \\
    \bottomrule
    \bottomrule
    \end{tabular}%
  \label{tab:quantized training}%
\end{table}%

\subsection{Extra Memory and Computation Cost of Back-propagation}\label{par: Extra Memory and Computation Cost of Back-propagation}


In this subsection, we analyze the extra memory and computation cost of BP. 

We consider a generic neural network with $L$ layers: 
\begin{equation}
\begin{aligned}
f(\mat{x} ; \boldsymbol{\theta}) & =f^{(L)}\left(f^{(L-1)}\left(\ldots f^{(0)}\left(\mat{x} ; \boldsymbol{\theta}^{(0)}\right) \ldots ; \boldsymbol{\theta}^{(L-1)}\right) ; \boldsymbol{\theta}^{(L)}\right).
\end{aligned}
\label{forward}
\end{equation}
Here $\bm{\theta}^{(i)}$ denotes the parameters of the $i$-th layer. 
The training process aims to find the optimal set of $\{\bm{\theta}^{*(i)}\}_{i=1}^L$ by minimizing an objective (loss) function $\mathcal{L}(f(\mat{x} ; \boldsymbol{\theta}), y)$. Here $f^{(i)}(\mat{a}^{(i)}; \boldsymbol{\theta}^{(i)})$ predicts the activation values $\mat{a}^{(i+1)}$ of the next layer based on its inputs $\mat{a}^{(i)}$. For instance, in a fully connected layer $\boldsymbol{\theta}^{(i)} =\{ \mat{W}^{(i)}; \mat{b}^{(i)} \}$, and $f^{(i)}$ performs the calculation
\begin{equation}
    \mat{a}^{(i+1)}=h^{(i)}(\mat{z}^{(i)}),\quad \mat{z}^{(i)}=\mat{W}^{(i)} \mat{a}^{(i)}+\mat{b}^{(i)}
\end{equation}
where $h^{(i)}$ is the nonlinear activation function.
The memory cost of an inference task consists of two parts: (1) the non-volatile storage (\textit{e.g.,} Flash) that stores the pre-trained model parameters $\{\bm{\theta}^{(i)} \in \mathbb{R}^{d_{\theta^{(i)}}} \}_{i=1}^L$, (2) run-time volatile memory (\textit{e.g.,} SRAM) that stores the intermediate computation results. The peak run-time memory at layer $i$ is approximately the summation $\mat{a}^{(i)}$, $\bm{\theta}^{(i)}$, and $\mat{a}^{(i+1)}$. 

{\bf Extra Memory of BP.}  In the BP process, given the gradients of the output activation $\nabla_{\mat{a}^{(i+1)}} \mathcal{L}$ at $i$-th layer, 
one needs to first back-propagate through the non-linear activation function:
\begin{equation}
    \nabla_{\mat{z}^{(i)}} \mathcal{L} = h'(\mat{z}^{(i)}) \nabla_{\mat{a}^{(i+1)}} \mathcal{L} 
\end{equation}
Then back-propagate through the linear transformation to compute the gradients of the input activation $\nabla_{\mat{a}^{(i)}} \mathcal{L}$ and the gradients of the parameters $\nabla_{\bm{\theta}^{(i)}} \mathcal{L}$: 
\begin{equation}
\nabla_{\mat{a}^{(i)}} \mathcal{L} = {\mat{W}^{(i)}}^T \cdot \nabla_{\mat{z}^{(i)}} \mathcal{L}, \quad \nabla_{\mat{W}^{(i)}} \mathcal{L} = \nabla_{\mat{z}^{(i)}} \mathcal{L} \cdot (\mathbf{a}^{(i)})^T, \quad \nabla_{\mat{b}^{(i)}} \mathcal{L}=\nabla_{\mat{z}^{(i)}} \mathcal{L}
\label{eq: backward}
\end{equation}
The result $\nabla_{\bm{\theta}^{(i)}} \mathcal{L} =\{ \nabla_{\mat{W}^{(i)}} \mathcal{L}, \nabla_{\mat{b}^{(i)}} \mathcal{L}\} $ is saved to update model parameters, and $\nabla_{\mat{a}^{(i)}} \mathcal{L}$ is propagated to the ($i$-1)-th layer. Throughout this paper, we use the term {\bf training memory} to denote run-time memory on an edge device.  
Apart from the memory to store the updated model parameters, we need the following additional memory in BP: 
\begin{itemize}
    \item {\bf Intermediate activation values} of all layers $\{\mat{a}^{(i)} \in \mathbb{R}^{B\times d_{a_i}} \}_{i=1}^L$ and $\{\mat{z}^{(i)} \in \mathbb{R}^{B\times d_{a_i}} \}_{i=1}^L$. Here $B$ denotes batch size and $d_{a_i}$ denotes the number of neurons in $i$-th layer.  
    \item {\bf Gradients} of model parameters $\{\nabla_{\bm{\theta}_{i}} {\bm{L}} \in \mathbb{R}^{d_{\theta_i}} \}_{i=1}^L$. Here $d_{\theta_i}$ denotes the dimension of parameters in $i$-th layer; 
    \item {\bf Optimizer state} (optional). Vanilla SGD \cite{bottou2010large} does not cost additional memory for the optimizer state. SGD with momentum and Adam optimizer \cite{kingma2014adam} costs $1 \times$ and $2 \times$ the size of all trainable parameters, respectively.
\end{itemize}
Previous on-device training frameworks explored techniques to alleviate the training memory bottleneck. 
For example, in convolutional neural networks, $\mat{a}^{(i)}$ need not be stored if the parameters in $i$-th layer are frozen as $\mat{a}^{(i)}$ is only involved in computing $\nabla_{\bm{\theta}^{(i)}} \mathcal{L}$ ~\citep{cai2020tinytl}.
For specific non-linear layers (\textit{e.g.,} ReLU~\cite{nair2010rectified} and other ReLU-styled), memory for $\mat{z}^{(i)}$ can be reduced by storing a binary mask representing whether the value is smaller than 0~\cite{cai2020tinytl}. However, such reductions are model-specific. Many operations (e.g., self-attention) involve $\mat{a}^{(i)}$ and many non-linear activations (\textit{e.g.,} GeLU~\cite{hendrycks2016gaussian}, Softmax, etc.) involve $\mat{z}^{(i)}$ in back-propagation, necessitating additional memory.
The memory-consuming nature of BP can only be alleviated, but not fully addressed. The inevitable extra memory to implement BP often prevents full-model training on resource-constrained edge devices. 

{\bf Extra Computation Graph.} The layer-by-layer computation in
Eq. \eqref{eq: backward} needs the knowledge of a computation graph. On cloud servers or personal computers with sufficient hardware resources and sophisticated operating system support, an AD package \cite{bolte2020mathematical} could automatically generate the computation graph. However, on edge devices running without an operating system (\textit{e.g.,} micro-controllers, FPGAs, ASICs, etc.), the computation graph and the corresponding hardware of Eq. \eqref{eq: backward} need to be hand-crafted and optimized for different models, tasks, and devices. This inevitably increase the design and manufacturing cost as well as the time to market.







\subsection{Zeroth-order Optimization}
\begin{figure}[t]
    \centering
    \includegraphics[width=0.9\linewidth]{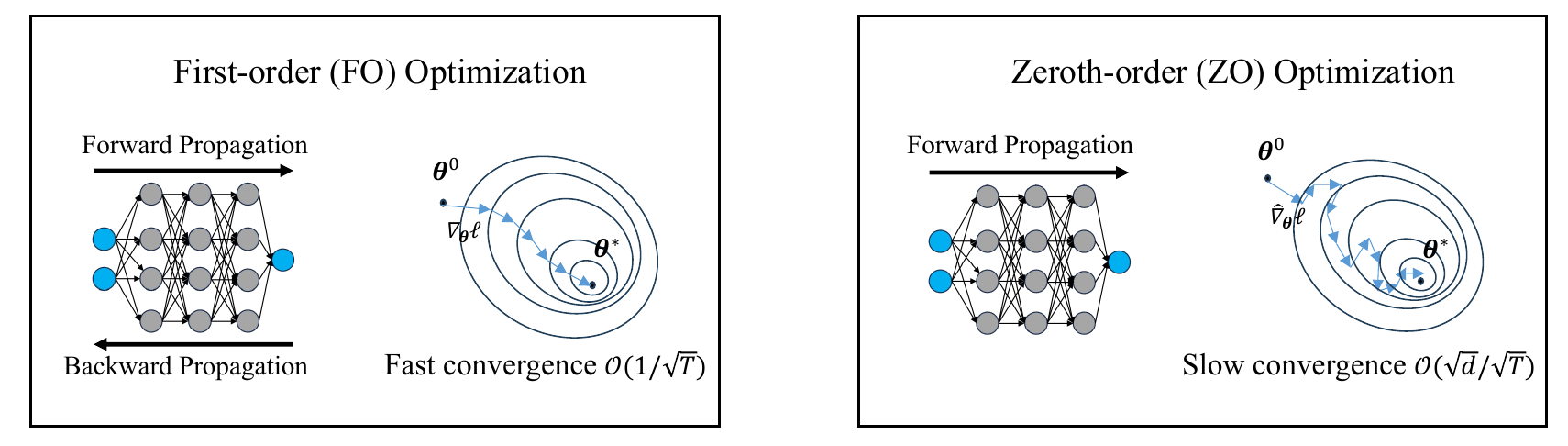}
    \caption{Comparison between first-order (FO) optimization and ZO optimization. FO optimization converges faster as it utilizes exact gradient from BP to update model parameters. ZO optimization, on the other hand, uses only forward function queries to estimate the gradients. ZO method converges more slowly due to the large variance of gradient estimation, but it is much more memory-efficient, since no extra computation graph is needed.}
    \label{fig:FO_ZO}
\end{figure}

As shown in Fig.~\ref{fig:FO_ZO}, zeroth-order (ZO) optimization~\cite{liu2020primer} uses only function queries, instead of exact gradient information, to solve optimization problems. In this work, we use the multi-point variant of randomized gradient estimator (RGE) \cite{liu2018zeroth, chen2023deepzero, malladi2023fine} as a ZO gradient estimator.


\begin{definition}[Randomized Gradient Estimator, RGE] For a finite-sum optimization problem $\text{min}_{\bm{\theta}\in \mathbb{R}^d} F(\bm{\theta}) = \frac{1}{n} \sum^n_{i=1} f_i (\bm{\theta})$, RGE estimates the gradients of $F$ with respect to variables $\bm{\theta} \in \mathbb{R}^d$ as:
\begin{equation}
   \hat{\nabla}_{\bm{\theta}}F(\bm{\theta})=\frac{1}{Q} \sum_{i=1}^Q 
    \frac{\left[F\left(\bm{\theta}+\mu \bm{\xi}_i \right)-F(\bm{\theta})\right] }{\mu} \bm{\xi}_i.
\label{eq: ZO-RGE}
\end{equation}
\end{definition}
Here $\{\bm{\xi}_i\}_{i=1}^Q$ are $Q$ \textit{i.i.d.} random perturbation vectors drawn from a zero-mean and unit-variance distribution (\textit{e.g.}, multivariate normal distribution $\mathcal{N}(\bm{0}, \bm{I})$ or multivariate uniform distribution $\mathcal{U}(\mathcal{S}(0,1)$ on a unit sphere centered at zero with a radius of one). $\mu >0$ is the sampling radius, which is typically small. RGE is an unbiased estimation of $\nabla_{\bm{\theta}}F_{\bm{\xi}}(\bm{\theta})$, here $F_{\bm{\xi}}(\bm{\theta})$ denotes the random smoothed version of $F(\bm{\theta})$. However, RGE is a biased estimation to $\nabla_{\bm{\theta}}F(\bm{\theta})$. With $\mu \rightarrow 0$, $\hat{\nabla}_{\bm{\theta}}F(\bm{\theta})$ is asymptotically unbiased to $\nabla_{\bm{\theta}}F(\bm{\theta})$ \cite{gao2022generalizing, chen2023deepzero}.



We utilize the corresponding stochastic gradient descent (SGD) \cite{bottou2010large} algorithm, ZO-SGD \cite{nesterov2017random, ghadimi2013stochastic} to update model parameters in the training process.

\begin{definition}[ZO-SGD] The variables $\bm{\theta}$ are iteratively updated as:
    \begin{equation}
    \bm{\theta}_t \leftarrow \bm{\theta}_{t-1}-\eta \hat{\nabla}_{\bm{\theta}}F(\bm{\theta})
    \label{eq: ZO-SGD}
\end{equation}
\end{definition}
Here the descent direction is computed using the ZO estimation rather than a BP method. 

\section{Poor Man's Training on MCU}

This section presents a completely BP-free training framework on MCU with tiny memory budget. Our goal is two-fold: (1) to enable ultra memory-efficient on-device full-model training, (2) to greatly simplify the design complexity of training hardware, making it as easy as inference hardware design. 

To achieve the above goals, we propose a quantized ZO optimization in Section \ref{par: All-quantized Training by Zeroth-Order Optimization} that employs only quantized inferences (forward evaluations) to optimize quantized model parameters. This allows us to reuse an inference hardware accelerator and convert it to a training engine with minimal changes. The ZO gradient estimation avoids the additional memory associated with storing activation values in BP. Here we employ the vanilla ZO-SGD [c.f. \eqref{eq: ZO-SGD}] without momentum to avoid the memory overhead caused by optimizer states. However, training a real-quantized model with ZO gradient estimation faces slow or even no convergence. This is caused by the dimension-dependent variance of ZO gradient estimation as well as the quantization error. We stabilize the algorithm by properly scaling the learning rate (c.f. Section~\ref{par: Learning Rate Scaling}) of each layer to mitigate the distorted gradient norms. Further, we employ a few dimension reduction methods to reduce the ZO gradient variance, including a combination of weight perturbation and node perturbation, as well as a BP-free sparse training method with task-adaptive trainable parameters selection.
These techniques combined enable a stable and accelerated BP-free training framework on an MCU. 

\subsection{Training Real Quantized Models via Quantized Zeroth-Order Optimization}\label{par: All-quantized Training by Zeroth-Order Optimization}
Given the resource constraints on an MCU, we consider a {\bf real-quantized} setting, where we only have access to the low-precision (integer) quantized and scaled representation of model parameters $\bm{\bar{\theta}}=\{ \mat{\bar{W}}, \mat{\bar{b}} \}$ and input/output activation values $\mat{\bar{a}}$. Let $\mathcal{L}(\bm{\theta}; \mathcal{X})$ be the empirical loss function, but we can only access via a sample-wise loss $\ell (\bm{\bar{\theta}}; \mat{x})$ evaluated at a data sample $\mat{x} \subset \mathcal{X}$. 
We estimate the gradient of quantized parameters $\nabla_{\bm{\bar{\theta}_{i}}} \mathcal{L}$ by a quantized ZO method and directly update the quantized model parameters.  

\subsubsection{Quantized ZO Gradient Estimation} Given a batch of data samples $\{\mat{x}_n\}_{n=1}^N \subset \mathcal{X}$, the quantized ZO randomized gradient estimation (quantized-RGE) is given as:

\begin{equation}
\hat{\nabla}_{\bm{\bar{\theta}}}\mathcal{L} = \frac{1}{NQ}  \sum_{n=1}^N \sum_{q=1}^Q \frac{\left[\ell \left(\bm{\bar{\theta}}+\mu \bm{\xi}_{n,q}; \mat{x}_n\right)-\ell (\bm{\bar{\theta}}; \mat{x}_n)\right]}{\mu} \bm{\xi}_{n,q}, \mat{x}_n \subset \mathcal{X}
\label{eq: quan-RGE}
\end{equation}
We jointly consider the double stochasticity of ZO-SGD, i.e., the stochasticity in sub-sampling training data and the stochasticity in random perturbation, by applying an independent set of perturbations $\{\{\bm{\xi}_q \}_{q=1}^Q\}_n$ to each training data sample $\mat{x}_n$.  
We directly add perturbation to the quantized model parameters $\bm{\bar{\theta}}\in \mathbb{R}^d$ with discrete integer values (\textit{e.g.,} $\{\bm{\bar{\theta}} \in \mathbb{Z} | -128\leq \bm{\bar{\theta}} \leq 127 \}$ for signed INT8 format). To ensure that the perturbed values are still in integer format, we sample the perturbation $\bm{\xi}_q$ from the Rademacher distribution of which the entries are integers +1 or -1 with equal probability. The perturbations sampled from a Rademacher distribution are zero-mean and unit-variance, ensuring that Eq. \eqref{eq: quan-RGE} is an unbiased gradient estimator as $\mu \rightarrow 0$ \cite{spall1992multivariate}. 

However, in a quantized setting, we must restrict the smoothing parameter $\mu$ to be the smallest integer (\textit{i.e.,} $\mu =1$). 
This leads to a biased gradient estimation. For SGD with a biased gradient estimator, the convergence rate depends on the mean squared error (MSE) of gradient estimation \cite{gao2022generalizing, demidovich2024guide}. The MSE of gradient estimation based on Eq. \eqref{eq: quan-RGE} derived by \cite{gao2022generalizing} is given as:

\begin{equation}
\mathbb{E}\left[ \| \hat{\nabla}_{\bm{\theta}} \mathcal{L} - \nabla_{\bm{\theta}} \mathcal{L} \|_2^2 \right] = \frac{d-1}{NQ}\left\|\nabla_{\bm{\theta}} \mathcal{L}(\bm{\theta})\right\|_2^2+\frac{d}{NQ} \operatorname{tr}\left(\operatorname{Var}_{\mat{x}}\left[\nabla_{\bm{\theta}} \ell (\bm{\theta}, \mat{x})\right]\right) + \mathcal{O}(\mu^2 d^2)
\label{eq: MSE error}
\end{equation}
On the right-hand side, the first term comes from the gradient estimation variance, the second term comes from the gradient variance from data sampling, and the last term is the remainder when $\mu$ does not go to 0. 
Reducing the MSE of the gradient estimation, especially the variance of gradient estimation that dominate MSE, provably improves the convergence speed \cite{gao2022generalizing}.

\subsubsection{Learning-rate Scaling}\label{par: Learning Rate Scaling} Directly updating quantized model parameters with the gradient estimator $\hat{\nabla}_{\bm{\bar{\theta}}}\mathcal{L}$ in Eq. \eqref{eq: quan-RGE} and using a global learning rate $\eta$ leads to slow or even no convergence.  Under this setting, we show that the SGD-style update is distorted by the high variance of gradient estimation and the quantization process. To address this issue, we incorporate two scaling methods to adjust the learning rate.

\begin{itemize}
    \item {\bf Gradient-Norm Scaling.} The high variance of a ZO gradient estimation distorts the gradient update. Lemma \ref{lemma: descent lemma}) derived by \cite{malladi2023fine} shows how the gradient norm influences the performance of an SGD optimizer.
    \begin{lemma}[Descent Lemma~\cite{malladi2023fine}]  Let $\mathcal{L}$ be l-smooth. For any unbiased gradient estimate $\hat{\nabla} \mathcal{L} \left(\boldsymbol{\theta}\right)$
        \begin{equation}
    \mathbb{E}\left[\mathcal{L}\left(\boldsymbol{\theta}_{t+1}\right) \mid \boldsymbol{\theta}_t\right]-\mathcal{L}\left(\boldsymbol{\theta}_t\right) \leq-\eta\left\|\nabla \mathcal{L}\left(\boldsymbol{\theta}_t\right)\right\|^2+\frac{1}{2} \eta^2 l \cdot \mathbb{E}\left[\left\|\hat{\boldsymbol{g}}\left(\boldsymbol{\theta}\right)\right\|^2\right]
    \end{equation}
    \label{lemma: descent lemma}
    \end{lemma}
    Lemma \ref{lemma: descent lemma} indicates that the largest permissible learning rate of ZO-SGD should be $\mathbb{E}\left[\| \hat{\nabla} \mathcal{L}_{\text{ZO-SGD}}  \|^2 \right] / \mathbb{E}\left[\| \hat{\nabla} \mathcal{L}_{\text{SGD}} \|^2 \right]$ times smaller than that of SGD to guarantee loss decrease.
    According to Eq. \eqref{eq: MSE error}, the squared norm of the ZO stochastic gradient estimation is approximately $(NQ+d-1)/NQ$ times larger than that of the FO stochastic gradient \cite{malladi2023fine}.
    To avoid complicated hyperparameter tuning of the learning rate according to different $N$, $Q$, or $d$, we propose to fix a global learning rate and scale it by $NQ/(NQ+d-1)$ at each step. 
    
    \item {\bf Quantization-Aware Scaling~\cite{lin2022device}.} The quantization process also distorts the SGD update. Let $\bm{\theta}$ denote the full-precision parameter, and $\bar{\bm{\theta}}=\bm{\theta} / s_{\bm{\theta}}$ denote its quantized (\textit{e.g.,} INT8) and scaled representation. Here $s \ll 1$. $\bar{\bm{\theta}}$ is $1/s_{\bm{\theta}}$ times larger than $\bm{\theta}$ in magnitude, while according to the chains rule, the gradient magnitude of quantized representation $\hat{\nabla}_{\bm{\bar{\theta}}}\mathcal{L}$ is $s_{\bm{\theta}}$ times smaller than that of $\hat{\nabla}_{\bm{\theta}}\mathcal{L}$. To ensure the update of a quantized parameter representation follows the expected update in its original scale, we follow \cite{lin2022device} to apply a quantization-aware scaling to the learning rate of each parameter. Specifically, we fix a global learning rate and divide the learning rate of each quantized parameter by the square of its scaling factor $s_{\bm{\theta}}$ (c.f. Appendix \ref{appendix:QAS} for details). Consequently, at step $t$ we update quantized model parameters as:
    \begin{equation}
        \bar{\bm{\theta}}^{t+1} \leftarrow \text{clip} \left( \bar{\bm{\theta}}^{t} -  \left\lceil \frac{NQ}{NQ+d-1} \frac{1}{s_{\bm{\theta}}^{2}} \cdot \eta \hat{\nabla}_{\bm{\bar{\theta}}}\mathcal{L} \right\rfloor, -Q_N, Q_P \right)
        \label{eq: quantized update}
    \end{equation}
    The update is rounded and clipped to maintain the updated parameters in its low-precision format. 
\end{itemize}

The convergence speed of ZO training depends on the gradient error \cite{gao2022generalizing, demidovich2024guide}, which is dominated by the gradient estimation variance [i.e., the first term in Eq. \eqref{eq: MSE error}].  
Despite various variance reduction methods for ZO optimization, extra memory is needed to save either the co-variance matrix \cite{shu2023zeroth}, hessian diagonal \cite{ye2018hessian}, or history perturbations \cite{cheng2019improving, liu2018zeroth, ji2019improved, di2024double}, making them unsuitable for edge devices with tiny memory budget. Ref. \cite{gao2022generalizing} proposed to find the optimal perturbation distribution that minimizes the MSE error, but it requires the shrinkage of the perturbation scale, making it infeasible for real-quantized graphs.
As the ZO gradient variance depends on the dimension $d$ of the optimization variables, we employ dimension-reduction methods to reduce the MSE error and improve the training convergence.

\subsection{Memory-Efficient Adaptive Weight/Node Perturbation}

\subsubsection{\bf Background: Dimension Reduction via Node Perturbation}

\begin{figure}
    \centering
    \includegraphics[width=0.9\linewidth]{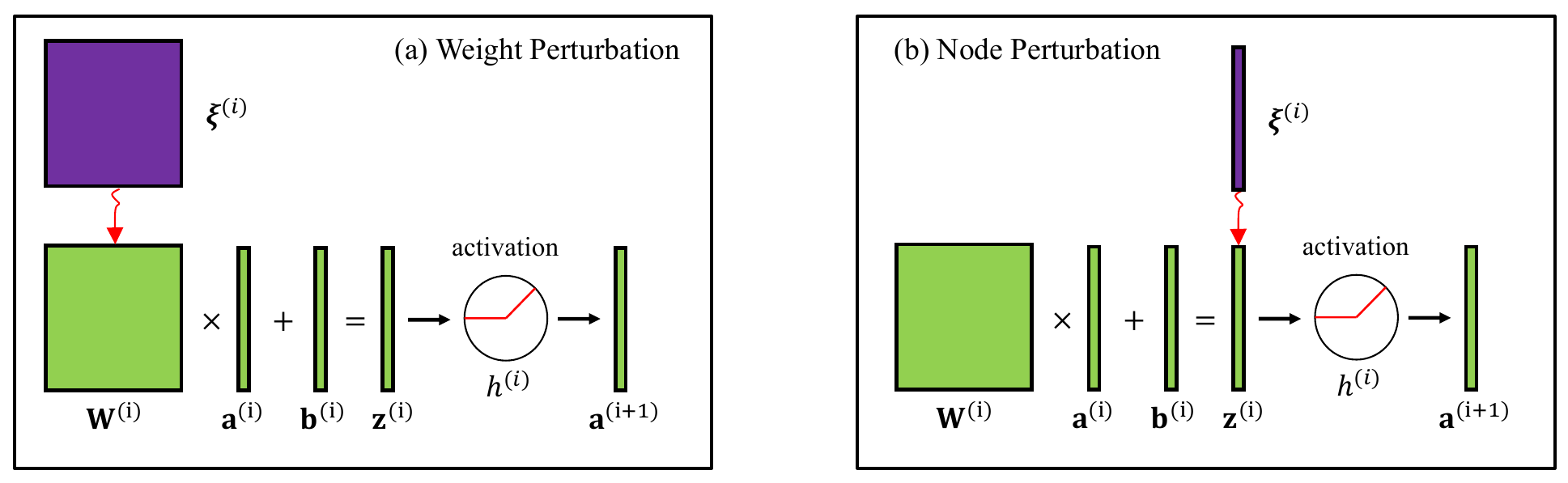}
    \caption{(a) Data flow of weight perturbation. (b) Data flow node perturbation. }
    \label{fig:modelwise perturbation}
\end{figure}


Eq. \eqref{eq: quan-RGE} perturbs the model parameters of every layer to obtain a ZO gradient. An alternative of obtaining the ZO gradient estimation for $\mat{W}_i$ and $\mat{b}_i$ is to firstly perturb the output nodes of linear transformation $\mat{z}^{(i)}=\mat{W}^{(i)} \mat{a}^{(i)}+\mat{b}^{(i)}$ and estimate the gradient of nodes $\hat{\nabla}_{\mat{z}^{(i)}} \mathcal{L}$, 
\begin{equation}
\hat{\nabla}_{\mat{z}^{(i)}}\mathcal{L} = \frac{1}{NQ} \sum_{n=1}^N \sum_{q=1}^Q \frac{\ell_q(\bar{\bm{\theta}}; \mat{x}_n) - \ell (\bar{\bm{\theta}}; \mat{x}_n) }{\mu} \bm{\xi}_{k,q,n}, \quad \mat{x}_n \subset \mathcal{X}.
\end{equation}
Here $\ell_q(\bar{\bm{\theta}}; \mat{x}_n)$ is the sample-wise loss associated with perturbing the linear transformation output $\mat{z}^{(i)}$ via $\mat{\bar{a}}^{(i+1)}_q= h^{(i)}(\mat{\bar{z}}^{(i)} + \mu \bm{\xi}^{(i)}_{q})$.
According to Eq. \eqref{eq: backward}, the gradient of the weights and biases is then obtained as \cite{ren2022scaling, dalm2023effective, hiratani2022stability}:
\begin{equation}
    \hat{\nabla}_{\mat{\bar{W}}^{(i)}} \mathcal{L} = \hat{\nabla}_{\mat{\bar{z}}^{(i)}} \mathcal{L} \cdot (\bar{\mat{a}}^{(i)})^T
    \quad \hat{\nabla}_{\mat{\bar{b}}^{(i)}} \mathcal{L}=\hat{\nabla}_{\mat{\bar{z}}^{(i)}} \mathcal{L}
    \label{eq: grad_a_to_grad_w}
\end{equation}
We term these two gradient estimation schemes as \textit{weight perturbation (WP)} and \textit{node perturbation (NP)}, respectively, which are illustrated in Fig. \ref{fig:modelwise perturbation}.
In cases where the activation dimension $d_a$ is smaller than weight dimension $d_w$, node perturbation benefits from a smaller gradient estimation variance of $\hat{\nabla}_{\mat{\bar{W}}_{i}} \mathcal{L}$.

\begin{figure}[t]
    \centering
    \includegraphics[width=0.9\linewidth]{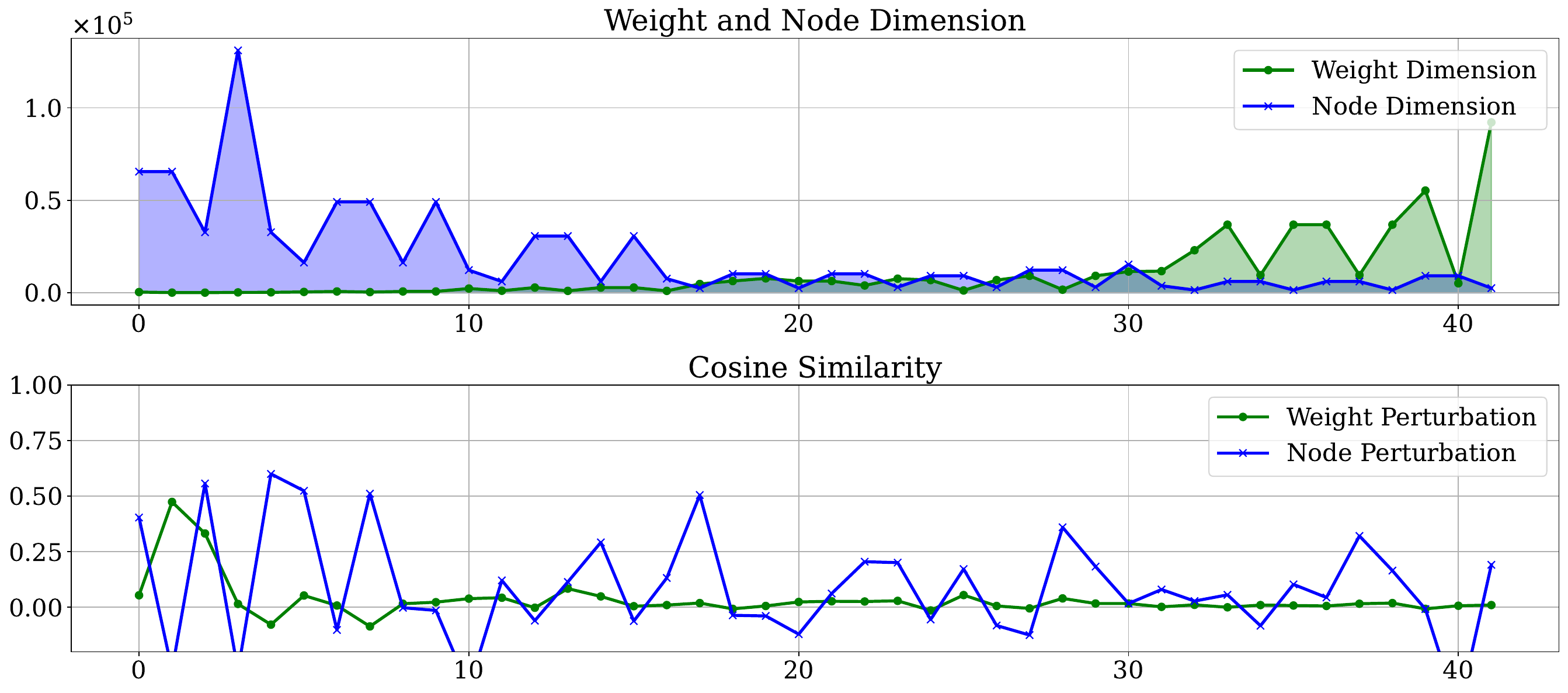}
    \caption{Top: The dimension of weights/nodes at each layer. Bottom: The cosine similarity between zeroth-order gradient estimation and the first-order gradient computed by back-propagation at each layer.}
    \label{fig:AP_WP_cos_sim}
\end{figure}

However, the superiority of node perturbation over weight perturbation could be hindered:
\begin{itemize}
    \item \textbf{Unbalanced Weight/Node Dimension:} Convolution neural networks have smaller weight dimensions and larger node dimensions in starting layers while having larger weight dimensions and smaller node dimensions in ending layers. We take a preliminary investigation into the gradient estimation variance at each layer. Fig. \ref{fig:AP_WP_cos_sim} shows the cosine similarity between the ZO gradient estimation and the FO gradient computed by BP of each layer. The larger cosine similarity indicates a better alignment with the true gradient, \textit{i.e.}, smaller gradient estimation MSE. Applying weight perturbation or activation perturbation across the whole model does not guarantee a better gradient estimation for all layers.
    \item \textbf{Inter-layer Feature Correlation:} Fig \ref{fig:AP_WP_cos_sim} also shows a zig-zag pattern in node perturbation. This is attributed to the simultaneous perturbation and update of all layers in vanilla node perturbation. 
    The perturbations added at each layer accumulate and get amplified to the following layers, which potentially introduces additional variance and affect the gradient estimation performance~\cite{hiratani2022stability, dalm2023effective}.
    \item \textbf{Memory Inefficiency:} Vanilla implementation of node perturbation has the same memory overhead as BP since the input activation values $\mat{a}_{i}$ of each layer need to be temporarily saved in the forward propagation. In contrast, by leveraging a pseudo-random number generator that could generate the same perturbation given the same random seed, weight perturbation needs storing only $2N+1$ scalars~\cite{malladi2023fine}.
\end{itemize}

\subsubsection{\bf Proposed Solution: Layer-Wise Memory-Efficient Adaptive Weight/Node Perturbation.} To fully leverage the benefit of node perturbation and improve the overall training performance, we propose the following techniques:
\begin{itemize}
    \item \textbf{Layer-wise Gradient Estimation.} We propose to leverage a layer-wise gradient estimation strategy. Specifically, at each step, we only add perturbation to, and estimate the gradients of the parameters in one single layer. We theoretically and experimentally show that layer-wise gradient estimation contributes to a lower gradient variance for each layer than model-wise gradient estimation.
Table \ref{tab:modelwise layerwise variance} summarizes the theoretical results. For simplicity, we consider a neural network with $L$ identical layers. The dimension of weight is $d_w$ and the dimension of output activation is $d_a$. The mini-batch (number of data samples) is $N$. We fix the computation cost (evaluated by the number of forward passes) of different methods for a fair comparison. The number of forward evaluations is $Q$ in model-wise perturbation and $Q/L$ for each layer in layer-wise perturbation.
Theoretically, the variance of layer-wise gradient estimation is just marginally smaller than that of model-wise gradient estimation, but this is true only if we neglect the possible cross-layer correlation between weights/nodes.
Empirically, Fig. \ref{fig:modelwise_layerwise_cos_sim} shows that layer-wise gradient estimation outperforms model-wise gradient estimation for both weight perturbation and node perturbation with a non-negligible gap.
This is attributed to the de-correlation between layers in a layer-wise gradient estimation method \cite{dalm2023effective}. 

\begin{table}[t]
  \centering
  \caption{Gradient estimation variance comparison between weight perturbation (WP) and node perturbation (NP) with their model-wise and layer-wise gradient estimation implementations.  $S$ is the squared gradient norm, and $V$ is the variance of stochastic gradient.}
    \begin{tabular}{c|ccc}
    \toprule
    \toprule
          & Computation & Model-wise Variance & Layer-wise Variance \\
    \midrule
    Weight Perturbation & $NQ$  & $\frac{Ld_w-1}{NQ} S+\frac{L}{NQ} V$ & $\frac{Ld_w-L}{NQ} S+\frac{L}{NQ} V$ \\
    Node Perturbation & $NQ$  & $\frac{Ld_a-1}{NQ} S+\frac{L}{NQ} V$ & $\frac{Ld_a-L}{NQ} S+\frac{L}{NQ} V$ \\
    \bottomrule
    \bottomrule
    \end{tabular}%
  \label{tab:modelwise layerwise variance}%
\end{table}%
\begin{figure}[t]
    \centering
    \includegraphics[width=0.9\linewidth]{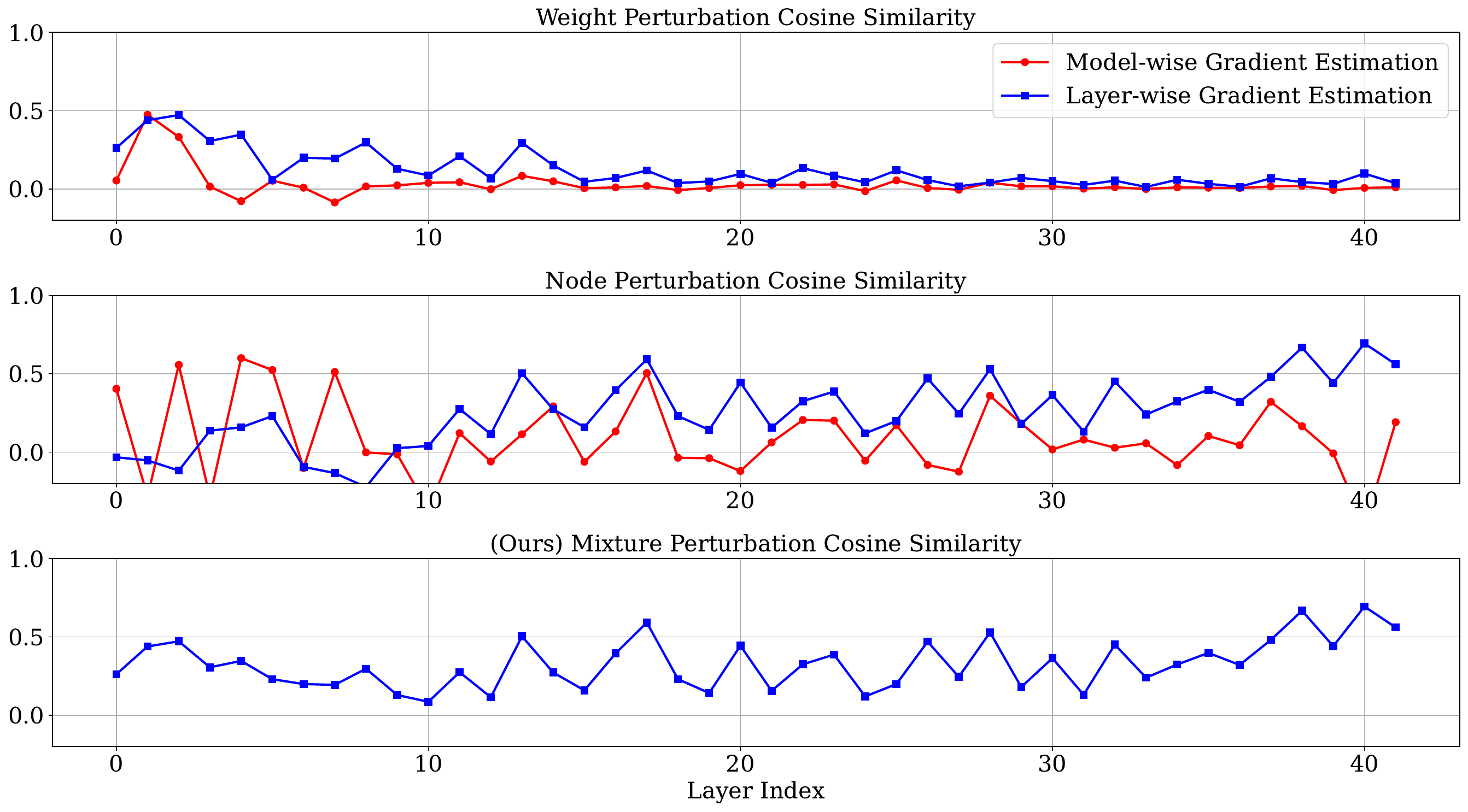}
    \caption{Quality of gradient estimation (measured via Cosine similarity) via weight perturbation (top), node perturbation (middle) and our mixture of weight and node perturbation (bottom). }
    \label{fig:modelwise_layerwise_cos_sim}
\end{figure}

    \item \textbf{Adaptive Weight Perturbation and Node Perturbation.} With a disentangled layer-wise gradient estimation scheme, we can minimize the gradient variance of each layer by applying weight perturbation for layer $i$ if $d_w^{(i)}<d_a^{(i)}$; otherwise, we apply node perturbation.  We can consistently get a lower gradient estimation variance regardless of the unbalanced weight and node dimension of different layers, as shown in Fig \ref{fig:modelwise_layerwise_cos_sim}. Note that the dimension $d$ in Eq. \eqref{eq: quantized update} is set as $d=d_w^{(i)}$ for layer-wise weight perturbation, and $d=d_a^{(i)}$ for layer-wise node perturbation, accordingly.
    \item \textbf{Memory-efficient Layer-wise Gradient Estimation.} We propose a memory-efficient implementation of node perturbation that minimizes the memory overhead. Different from BP, we only need $\mat{a}_{i}$ for the gradient estimation of $i$-th layer in node perturbation. With the aforementioned layer-wise gradient estimation, we estimate $\hat{\nabla}_{\mat{a}_{i+1}} {\mathcal{L}}$, compute the gradients of weights $\hat{\nabla}_{\mat{W}_{i}} {\mathcal{L}}$, and update $\mat{W}_i$ instantly after the forward computation of the $i$-th layer. In this way, $\mat{a}_{i}$ needs not be temporarily stored. Inspired by \cite{malladi2023fine}, we only store the random seed that generates the perturbation instead of the whole perturbation matrix. When regenerating $\hat{\nabla}_{\mat{a}_{i+1}} {\mathcal{L}}$, we could reuse the same memory that temporarily stored $\mat{a}_{i+1}$. Note that the old values of $\mat{W}_i^{t-1}$ should be stored (via reusing the memory of $\hat{\nabla}_{\mat{W}_{i}} {\mathcal{L}}$ by in-place swap) when $\mat{W}_i$ is updated. In this way, we could re-compute $\mat{a}_{i+1}^{t-1}$, which guarantees the activations of the following layers are still in the $(t-1)$-th step. This ensures a \textit{global} update (the updates of all layers are synchronized), compared with a \textit{local} update like layer-wise block-coordinate descent \cite{cai2021zeroth} where different blocks (layers) are updated sequentially. The theoretical study \cite{baldi2016theory} shows that local updates can generalize worse than global updates.
    The peak training memory comparison between inference-only, BP-based training, memory-efficient weight perturbation, and our memory-efficient node perturbation is provided in Table \ref{tab:modelwise layerwise memory}. The peak memory of our method is slightly larger than that of memory-efficient weight perturbation but is much smaller than that of BP.
    \begin{table}[t]
  \centering
  \caption{Memory cost comparison.  FWD denotes forward propagation, and BWD denotes backward propagation.}
\begin{tabular}{c|ccc}
\toprule
\toprule
      & Computation & Peak Memory (Vanilla) & Peak Memory (Efficient) \\
\midrule
Inference Only & $N\times$ FWD & $N(2d_a+d_w)$ & / \\
Weight Perturbation & $NQ\times$ FWD & $Ld_w$ & $LQ$ \\
Node Perturbation & $NQ\times$ FWD & $NLd_a + Ld_w$ & $NLQ+d_w$ \\
Back-propagation & $N\times$ (FWD + BWD) & $NLd_a + Ld_w$ & / \\
\bottomrule
\bottomrule
\end{tabular}%
  \label{tab:modelwise layerwise memory}%
\end{table}%

\end{itemize}

Algorithm \ref{alg: Me} in the appendix gives the pseudo-code for the whole gradient estimation and update procedure.

\subsection{Further Dimension Reduction via Sparse Training} \label{par: Sparse Training}

We propose to further reduce the dimension by integrating sparse training with our BP-free training framework.
Sparse training identifies and updates only the ``most important" parameters while freezing others, significantly reducing the number of trainable parameters. By incorporating sparse training with BP-free training, we expect to effectively reduce the variance in ZO gradient estimation. While previous work has shown that properly selected trainable parameters can match or surpass full-model training in fine-tuning pre-trained models~\cite{lee2022surgical, lin2022device, kwon2023tinytrain}, existing methods are ill-suited for resource-constrained on-device training on MCUs for two main reasons:
\begin{itemize}
    \item {\bf Incompatibility with on-device training.} Current methods either involve model training \cite{lin2022device, frankle2018lottery, zhang2022advancing} or evaluating gradient-based ``importance" metric \cite{kwon2023tinytrain,huang2023towards,sung2021training} to determine the ``effective" parameters for each target dataset. 
    Both necessitates a full BP graph which is not supported on MCUs. 
    \item {\bf Additional Memory Overhead.} Current methods use a fine selection granularity (\textit{e.g.,} parameter selection)~\cite{chen2023deepzero}. It introduces additional memory as large as a model copy to maintain an importance score for all parameters.
\end{itemize}

\subsubsection{\bf Proposed BP-free Sparse Training} \label{par: Layer Selection} 
To address these challenges, we propose a BP-free sparse training method compatible with MCUs' constrained memory resources. Our approach consists of two steps: BP-free trainable parameters selection and BP-free sparse training.
In the first step, we employ a brute-force, yet effective BP-free selection metric. 
We follow the surgical fine-tuning settings in ~\cite{lee2022surgical} to split the model into four non-overlapping blocks. We denote them as "Block 1", "Block 2", etc. in the order of input to output.
For a new target dataset, we conduct a BP-free test training (1 epoch) for each block. The block achieving the highest accuracy gain on a held-out subset of the training data is selected.
Then, we proceed with BP-free sparse training, where we only estimate the gradients of and update the parameters in the selected block, while keeping other blocks frozen. Our solution comes with the following advantages:
\begin{itemize}
    \item {\bf Easy Implementation.} BP-free gradient computation shares the same computation graph regardless of which parameters are selected for training. After deploying the BP-free training framework on MCUs, changing the selection of trainable parameters is purely a \textbf{software} effort. 
    \item {\bf MCU-compatible Overhead.} The coarse selection granularity introduces a minimal memory overhead (several scalars) and minimal computation overhead for selection (4 epochs).
    \item {\bf On-device Task-adaptive Selection.}
    The MCU-compatible overhead allows fully on-device evaluation of the selection process and adapting to different datasets without external computation.
\end{itemize}
This approach accelerates training convergence and achieves shorter end-to-end training times. As shown in Table \ref{tab:cifar-accuracy}, sparse BP-free training can match or even surpass full-model BP-based training (FullTrain) in certain cases.

\subsection{MCU Implementation Details}


In this section, we describe the implementation details of our BP-free training on MCUs. We deploy full-model training on STM32H7B3 MCU (1184-KB SRAM including 1024-KB user SRAM and 2-MB Flash) and deploy sparse training on STM32F746 MCU (320-KB SRAM including 256-KB user SRAM and 1-MB Flash). 

\begin{figure}[t]
    \centering
    \includegraphics[width=0.6\linewidth]{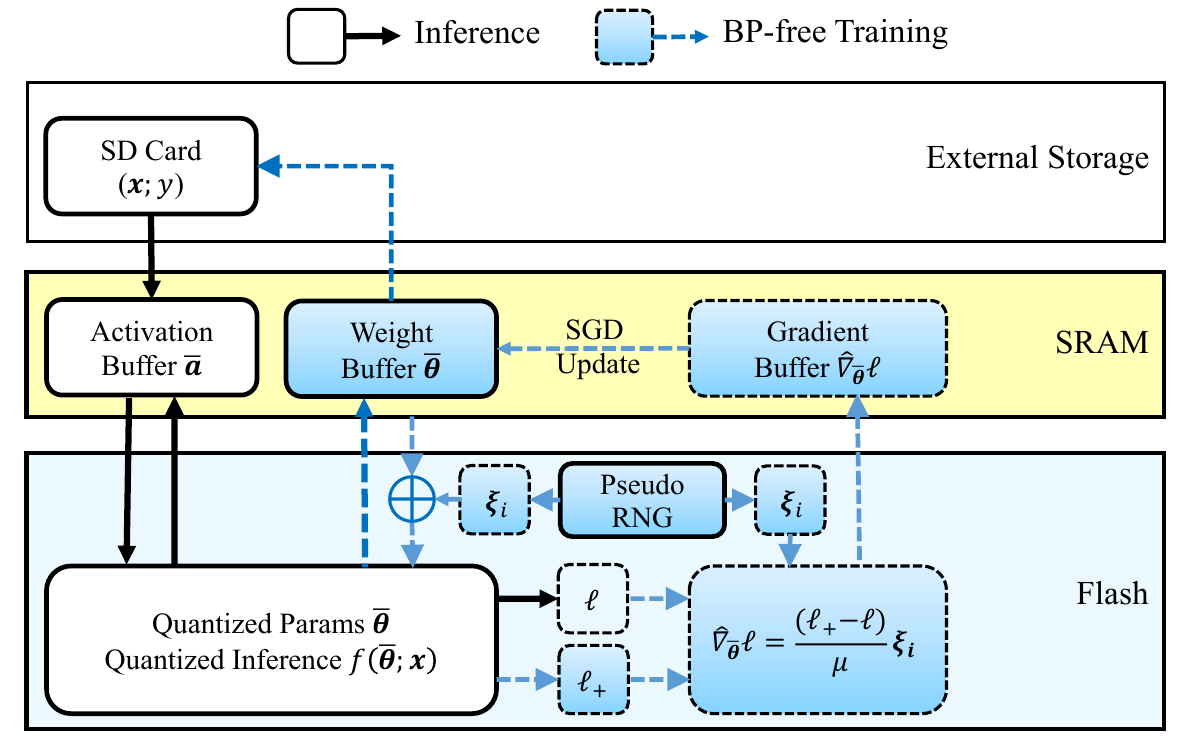}
    \caption{Overview of BP-free training framework on MCU. }
    \label{fig:MCU_diagram}
\end{figure}

\subsubsection{Overall Framework.} Fig. \ref{fig:MCU_diagram} shows the overall training framework. A few key points are summarized below:
\begin{itemize}
\item {\bf Quantized Inference Engine.} We employ TinyEngine \cite{lin2022device, zhu2023pockengine} to deploy tailored neural network model on the MCU. As an example, we consider the tailored CNN model MCUNet-in1 \cite{lin2020mcunet} with 0.46M parameters. The model parameter memory usage is 478 KB, and the peak inference run-time memory (SRAM) usage is 190 KB. The inference latency is measured as 239 ms using a single batch size with input resolution 128$\times$128.

\item {\bf Run-time Memory:} 
The pre-trained parameters are hard-coded in the read-only Flash. We first store the trainable parameters in the weight buffer on SRAM. 
The activation buffer temporarily stores the intermediate activation between layers until they are not needed for future computation.
The gradient buffer temporarily stores the gradients, and it is released once the parameter update is complete. In case we use gradient accumulation over multiple training data samples, the gradient buffer is kept between different forward (inference) runs.

\item {\bf External Memory:} We store the training data in an SD card. During training, no intermediate values are saved in the external memory. Once the training is complete, we store the checkpoint of updated model parameters in the SD card.

\item {\bf Gradient Estimation and Parameter Update.} We deploy the memory-efficient layer-wise gradient estimation with adaptive weight/node perturbation. The step-by-step implementation is summarized in Algorithm \ref{alg: Me}.
\end{itemize}

\subsubsection{In-Place Random Perturbation Generation} We generate the quantized perturbation vectors based on a Rademacher distribution and using the XORShift pseudo-random number generator \cite{panneton2005xorshift, marsaglia2003xorshift}. 
XORShift uses bitwise exclusive OR (XOR) and bit shifts (left and right) to produce a sequence of pseudo-random numbers efficiently in an iterative way: 
\[
\text{state} \leftarrow \text{state} \oplus (\text{state} \ll a)
\]
\[
\text{state} \leftarrow \text{state} \oplus (\text{state} \gg b)
\]
\[
\text{state} \leftarrow \text{state} \oplus (\text{state} \ll c)
\]
Here $a$, $b$, and $c$ are constants; $\oplus$ denotes the bitwise XOR operator; $\leftarrow$ and $\rightarrow$ denote left and right bit shifts, respectively. These operations ensure sufficient mixing of bits in the state, generating pseudo-random numbers with desirable statistical properties. In practice, the constants are commonly chosen as $a=13$, $b=17$, and $c=5$. We use the following steps to generate a sequence of random numbers following the Rademacher distribution:
\begin{enumerate}
    \item \textbf{Initialize the seed.} We set the seed for the XORShift generator, which will serve as the initial state.
    \item \textbf{Update the state.} We apply the XORShift recurrence relations to the current state to generate a new pseudo-random number.
    \item \textbf{Generate Rademacher-distributed samples.} The pseudo-random number is transformed into a Rademacher-distributed value based on the least significant bit (LSB). If the LSB is $1$, return $-1$; otherwise, return $+1$.
\end{enumerate}
The random perturbations only need to be generated on-demand and added to the weights/activations in-place. The same random perturabtion can be re-generated using the same initial seed. Therefore, we do not need a buffer to temporarily store the random perturbations.

\section{Experimental Results}

\subsection{Experiment Setup}\label{par:exp-setups}

{\bf Datasets.} We evaluate various training methods on multiple image classification datasets of two types:
\begin{itemize}
    \item {\bf Image Corruption.} We evaluate our BP-free training framework on a widely used out-of-distribution image corruption benchmarks \textbf{CIFAR-10-C} \cite{hendrycks2019benchmarking}. The task is to classify images from the target datasets, which consist of images corrupted by different kinds of corruptions un-seen in the source pre-training dataset with 5 severity levels. We run experiments over 10 of the corruptions (gaussian noise, impulse noise, shot noise, fog, frost, snow, defocus blur, elastic transform, brightness, contrast, and defocus blur). We tune on a training dataset with 1000 images and evaluate on a held-out test dataset from each of the corruptions. 
    \item {\bf Fine-grained Vision Classification (FGVC).} Following \textit{MCUTrain} \cite{lin2022device} and \textit{TinyTrain} \cite{kwon2023tinytrain}, we also test our BP-free training framework on several Fine-grained Vision Classification (FGVC) datasets including Cars \cite{krause20133d}, CUB \cite{welinder2010caltech}, Flowers \cite{nilsback2008automated}, Food \cite{bossard2014food}, and Pets \cite{parkhi2012cats}. The training datasets contain approximately 30-50 samples of each class.
\end{itemize}

\textbf{Model Configurations:}
We employ the MCU-compatible convolutional neural network (CNN) model \textit{MCUNet-in1}~\cite{lin2020mcunet} with 22.5M MACs and 0.46M parameters. The model is pre-trained on ImageNet-1k~\cite{deng2009imagenet} and quantized to the INT8 format by post-training quantization~\cite{jacob2018quantization}. The batch normalization~\cite{ioffe2015batch} layers are fused into the adjacent convolution layer. Only quantized model parameters and quantized inference computation graphs are implemented on the MCU. Unless otherwise stated, all models used in this paper are in the INT8 format.

\textbf{Training Configurations:}
We use a resolution of 128$\times$128 following~\cite{lin2022device} and gradient accumulation with 100 steps for all datasets, models, and baselines for a fair comparison. We apply vanilla SGD~\cite{bottou2010large} / ZO-SGD~\cite{nesterov2017random} without momentum as the optimizer to avoid additional memory cost of optimizer states. No weight decay was applied.
For experiments on the CIFAR-10-C datasets, the MCUNet model pre-trained on ImageNet-1k is transferred to the CIFAR-10~\cite{krizhevsky2009learning} dataset with 90.17\% test accuracy. In our BP-free training and the BP-based training baselines, we train the model for 50 epochs with an initial learning rate of 0.01 and a cosine decay following the setups in~\cite{lee2022surgical}.
For experiments on FGVC datasets, we train the model for 50 epochs following~\cite{lin2022device} in both BP-based and our BP-free methods. The initial learning rate is set as 0.1, and a cosine decay method is used.
In the BP-free training scenarios, the smoothing factor is set as $\mu=1$, the query budget is $Q=100$ for each layer. The trainable parameter selection is evaluated only once at the beginning of training.
To obtain experiment results of multiple downstream datasets faster, we simulate the training process on GPU with batch size 100. As our model does not have any batch-dependent operations (e.g., batch normalization~\cite{ioffe2015batch}), training with batch size 100 is equivalent to training with batch size 1 along with 100 steps of gradient accumulations, with learning rate scaled accordingly. 
For node perturbation, i.i.d. perturbations are simultaneously applied to the pre-activation vector in a mini-batch at each layer. For weight perturbation, the i.i.d. perturbations are shared across a mini-batch.

\subsection{Algorithmic Performance.}

\begin{figure}[t]
\centering
\begin{minipage}[t]{0.45\textwidth}
    \centering
    \subfigure{
        \includegraphics[width=\textwidth]{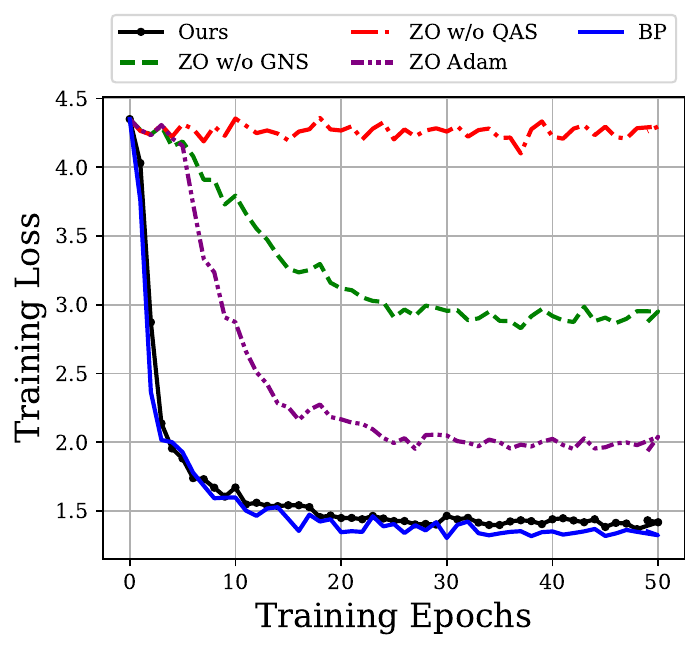}
        \label{fig:abla_loss_scaling}
    }
\end{minipage}
\begin{minipage}[t]{0.45\textwidth}
    \centering
    \subfigure{
        \includegraphics[width=\textwidth]{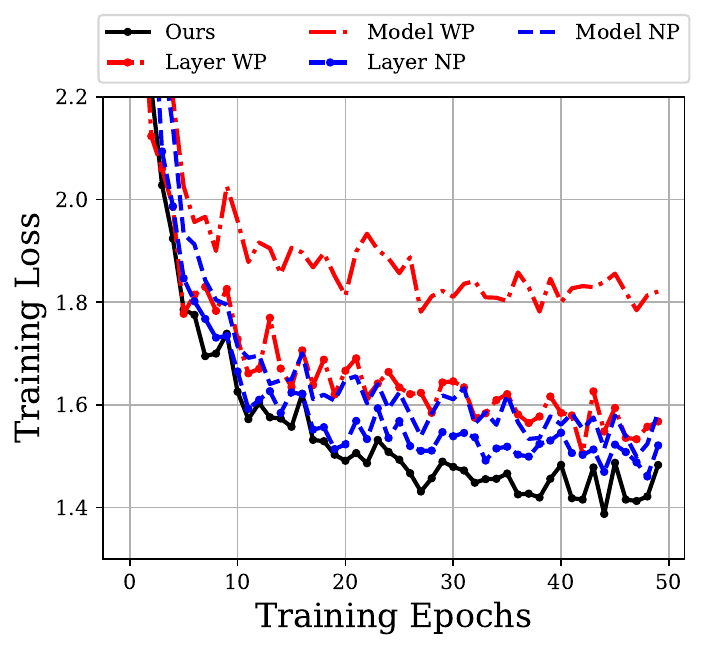}
        \label{fig:abla_loss_recipe}
    }
\end{minipage}
\quad
\caption{Left: Training loss curves w/ and w/o learning-rate scaling applied. QAS: quantization-aware scaling. GNS: gradient-norm scaling. Right: Training loss curves with different zeroth-order gradient estimation methods applied. We select the best learning rate for each method.
}
\label{fig:abla_loss}
\end{figure}

\subsubsection{Effectiveness of Various ZO Optimization Improvement Techniques.} We first validate the effectiveness of the key convergence improvement techniques used in our method. To simplify the comparison, we consider fine-tuning the MCUNet on the CIFAR-10-C with a Gaussian noise corruption at severity 5 and fix the trainable parameters as the weight matrices and biases of four point-wise convolution layers in block 1. 

\begin{itemize}[leftmargin=*]
\item {\bf Effectiveness of Learning Rate Scaling.} 
The training curves with and without learning rate scaling are provided in Figure \ref{fig:abla_loss} (Left). 
We apply layer-wise activation perturbation for the ZO gradient estimation.
Without gradient-norm scaling or quantization-aware scaling, the training converges slowly or even cannot converge. Adaptive learning rate methods like Adam \cite{kingma2014adam} cannot fully address this challenge while costing $3\times$ memory to save the optimizer states. With both scaling methods applied, our BP-free training follows a similar training dynamics as full-precision BP-based training without using any extra memory. 

\item {\bf Effectiveness of the Layer-Wise Mixture of Weight and Node Perturbations.} We further compare the effectiveness of different perturbation methods. The first two layers have larger node dimensions, while the other two layers have larger weight dimensions. The training results are provided in Figure \ref{fig:abla_loss} (Right). For a fair comparison, we consider the same computation budget (measured by the number of forward passes) for gradient estimation at each step. Layer-wise gradient estimation outperforms model-wise gradient estimation in both weight and node perturbations, showing that dis-entangling the gradient estimation layer-by-layer improves convergence. Our layer-wise gradient estimation with adaptive weight/node perturbation outperforms all other methods and achieves the best training convergence.

\item {\bf Effectiveness of Task-Adaptive Sparse Training.} Table \ref{tab:cifar_layer_selection} compares the average testing accuracy across various corruption types after training different blocks. Since different corruption types introduce varying distribution shifts from the pre-trained data, they benefit from fine-tuning different layers. A fixed selection of trainable blocks cannot guarantee optimal performance across all corruption types. Therefore, a task-adaptive selection is crucial, and our on-device selection metric is able to identify the most effective block at the onset of training. Consequently, our adaptive approach consistently outperforms all fixed selection methods.

\end{itemize}



\begin{table}[htbp]
  \centering
  \caption{Average test accuracy comparison of training a fixed selection of layers for all corruptions and applying the task-adaptive layer selection for different corruptions.}
    \begin{tabular}{cc|cccc}
    \toprule
    \toprule
          &       & noise & blur  & weather & digital \\
    \midrule
    \multicolumn{1}{c}{} & Block 1 & 55.17 & 78.91 & 79.25 & 75.76 \\
    \multicolumn{1}{c}{Fixed} & Block 2 & \textbf{65.33} & 80.31 & 81.33 & 77.45 \\
    \multicolumn{1}{c}{Selection} & Block 3 & 59.75 & 79.51 & 81.33 & 79.32 \\
    \multicolumn{1}{c}{} & Block 4 & 57.11 & 78.56 & 79.31 & 77.11 \\
    \multicolumn{2}{c|}{Adative Selection} & \textbf{65.33} & \textbf{80.91} & \textbf{82.00} & \textbf{80.56} \\
    \bottomrule
    \bottomrule
    \end{tabular}%
    
  \label{tab:cifar_layer_selection}%
\end{table}%

\subsubsection{Performance Comparison with Other BP-free Training Methods} Next, we compare our proposed methods with two state-of-the-art BP-free training baselines: \textit{MeZO} \cite{malladi2023fine} and \textit{DeepZero} \cite{chen2023deepzero}. 
\textit{MeZO} \cite{malladi2023fine} uses RGE with memory-efficient model-wise weight perturbation as the gradient estimator. All trainable parameters are perturbed and updated at the same time.
\textit{DeepZero} \cite{chen2023deepzero} uses sparsity-assisted coordinate-wise gradient estimator (CGE). Each single parameter (coordinate) is perturbed sequentially. Note that the proposed gradient sparsity pattern exploration of \textit{DeepZero} needs additional memory as large as three model copies, which is prohibitive on tiny edge devices. Therefore, we only apply a random gradient sparsity.
Table \ref{tab:cifar-bp-free} shows the model adaptation results. The main results are summarized below:
\begin{itemize}[leftmargin=*]
    \item {\bf MeZO~\cite{malladi2023fine}.} With the same same total number of forward evaluations, \textit{MeZO} attains the least performance improvement.  MeZO fails to converge due to 1) lack of proper learning-rate scaling, 2) large variance of model-wise ZO gradient estimation with weight perturbations, and 3) the general behaviour learned in pre-training (\textit{e.g.,} feature extracting, classifier) being corrupted in ZO fine-tuning. In summary, MeZO is not an effective generic BP-free training method. We would like to remark that MeZO still fails to converge well even if our gradient scaling method is used. 
    \item{\bf DeepZero~\cite{chen2023deepzero}.} \textit{DeepZero} attains low-variance ZO gradient by coordinate-wise gradient estimation, at the cost of two orders-of-magnitude more forward evaluations per iteration. \textit{DeepZero} still fails to match the testing accuracy of our method. We posit that \textit{DeepZero}'s parameter-wise sparsity pattern scheme is less effective in detecting the important parameters for image corruption datasets.
    \item {\bf Our Method.} Our proposed BP-free training method achieves the best efficiency and accuracy compared with the above baseline methods.
\end{itemize}
We further point out that DeepZero and MeZO still cannot achieve the same performance as our method, even if these two baseline methods are improved by applying the same trainable parameter selection, as shown in Table~\ref{tab:cifar-bp-free-layer-selection}. 

\begin{table}[t]
  \centering
  \caption{Comparison between our BP-free training and other state-of-the-art BP-free training baselines.}
    \begin{tabular}{c|cccc}
    \toprule
    \toprule
          & iter. forwards & epochs & total forwards & test accuracy (\%) \\
    \midrule
    MeZO~\cite{malladi2023fine}  & 2     & 10000 & \textbf{2.00E+07} & 18.33 \\
    DeepZero~\cite{chen2023deepzero} & 49457 & 50    & 2.47E+09 & 52.78 \\
    Ours  & 400   & 50    & \textbf{2.00E+07} & \textbf{66.78} \\
    \bottomrule
    \bottomrule
    \end{tabular}%
  \label{tab:cifar-bp-free}%
\end{table}%

\begin{table}[htbp]
  \centering
  \caption{Comparison between our BP-free training and other state-of-the-art BP-free training baselines when applied the same trainable parameter selection. DeepZero-0 denotes dense training, and DeepZero-0.9 means training with a 90\% gradient sparsity.}
    \begin{tabular}{c|cccc}
    \toprule
    \toprule
          & iter. forwards & epochs & total forwards & test accuracy (\%) \\
    \midrule
    MeZO~\cite{malladi2023fine}  & 2     & 10000 & \textbf{2.00E+07} & 40.56 \\
    DeepZero-0~\cite{chen2023deepzero} & 18560 & 50    & 9.28E+08 & 60.67 \\
    DeepZero-0.9~\cite{chen2023deepzero} & 1856  & 50    & 9.28E+07 & 54.33 \\
    Ours  & 400   & 50    & \textbf{2.00E+07} & \textbf{66.78} \\
    \bottomrule
    \bottomrule
    \end{tabular}%
    
  \label{tab:cifar-bp-free-layer-selection}%
\end{table}%

\subsection{On-Device Training Results}
Now we implement our method on the MCU, and compare its performance with existing on-device training methods on two benchmarks: CIFAR-10-C datasets and FGVC datasets. 

{\bf Baselines.} We compare our method with the following four on-device back-propagation training baselines: (1) \textit{No Adaptation} does not perform any on-device training. (2) \textit{FullTrain} fine-tunes the entire model. (3) \textit{MCUTrain} is the state-of-the-art (SOTA) method for training under an extremely low memory budget (\textit{e.g.}, MCU with 256-KB SRAM), which statically determines the set of layers/channels to update based on \textit{offline} evolutionary search on the \textit{cloud} servers before MCU deployment, and updates the selected layers/channels on MCU. (4) \textit{TinyTrain \cite{kwon2023tinytrain}} is the most recent SOTA on-device training method that employs \textit{on-device} layer/channel selection. For \textit{MCUTrain} and \textit{TinyTrain}, we select the best layer selection setting of each method that could fit into the memory budget of an MCU with 256-KB SRAM.

{\bf Performance Evaluation.} Through our experiments, we evaluate the following performance metrics:
\begin{itemize}
\item {\bf Memory Cost.}\label{par: exp_memory_analysis}
We profile \textbf{analytic memory} to reflect the memory cost of different training methods. We count the memory required for training, including the peak memory of inference, trainable parameters, and the extra memory for gradient computation and parameter update. For training with back-propagation, the extra memory includes the saved intermediate activation, and the gradients of weights. For back-propagation-free training, the extra memory includes loss values, random seeds, and a buffer needed to save the largest gradients of weights (only for node perturbation). The analytic memory determines whether a can be deployed under the memory budget. We then measure \textbf{on-device memory} to measure the actual memory cost for MCU-deployable training methods.

\item {\bf Computation Analysis.}\label{par: exp_computation_analysis}
We count the number of MACs to reflect the computation cost of different training methods. For each training iteration, the computation consists of 1 forward propagation, gradient computation (by back-propagation or zeroth-order gradient estimation), and parameter update. The analytic computation cost is implementation-agnostic.

\item {\bf End-to-end Training Time.}
We measure the end-to-end training time as per iteration latency times the number of iterations.
The per-iteration latency depends on the implementation backends. We consider a general TFLite Micro \cite{lai1801efficient} inference backend and an MCU domain-specific inference backend TinyEngine \cite{lin2022device, zhu2023pockengine}. Note that TFLite Micro does not support training, TinyEngine has support for sparse training but the update space is constrained within some ending layers. We used the kernel implementations to profile the \textbf{projected per-iteration latency} to perform full-model back-propagation. 
We then measure the actual \textbf{on-device per-iteration latency} for MCU-deployable training methods.
\end{itemize}

\subsubsection{Results on The CIFAR-10-C Dataset}

\begin{table}[tbp]
\centering
\caption{Test Accuracy of CIFAR-10-C on 12 representative corruptions at severity level 5. The training settings that could be implemented on an MCU with 256 KB SRAM are \underline{underlined}. Other training settings are out-of-memory and only listed for comparison.}
\resizebox{\linewidth}{!}{
    \begin{tabular}{cc|cccc|cccc|cc|cc}
    \toprule
    \toprule
          &       & \multicolumn{4}{c|}{Noise}    & \multicolumn{4}{c|}{Blur}     & \multicolumn{2}{c|}{Weather} & \multicolumn{2}{c}{Digital} \\
    \cmidrule{3-14}Model & Method & Gauss. & Shot  & Impul. & Speckle & Gauss. & Defoc. & Motion & Zoom  & Fog   & Brit. & Contr. & Elas. \\
    \midrule
    \multicolumn{2}{c|}{No Adaptation} & 17    & 21    & 25    & 27    & 72    & 76    & 67    & 80    & 73    & 83    & 36    & 71 \\
    \midrule
    FP32  & FullTrain & 73.89 & 72.44 & 76.78 & 78.44 & 77.89 & 81.00 & 78.44 & 80.78 & 82.67 & 86.67 & 86.33 & 76.44 \\
    \midrule
          & FullTrain & 44.33 & 37.78 & 41.55 & 54.00 & 44.11 & 65.11 & 42.78 & 55.57 & 34.22 & 73.00 & 20.27 & 77.78 \\
    INT8  & \underline{MCUTrain~\cite{lin2020mcunet}} & 61.77 & 63.66 & 55.55 & 65.33 & \textbf{81.55} & 83.44 & \textbf{82.44} & \textbf{87.33} & 76.22 & \textbf{89.22} & \textbf{71.77} & \textbf{81.11} \\
          & \underline{BP-free (Ours)} & \textbf{66.78} & \textbf{66.22} & \textbf{61.00} & \textbf{67.33} & 80.11 & \textbf{83.66} & 81.00 & 86.33 & \textbf{78.33} & 88.22 & 69.78 & 79.56 \\
    \bottomrule
    \bottomrule
    \end{tabular}%

}
  \label{tab:cifar-accuracy}%
\end{table}%
Here we consider on-MCU training to adapt a pre-trained model to the CIFAR-10 dataset with various levels of input image corruptions.
\begin{itemize}[leftmargin=*]
    \item {\bf Accuracy.} Table \ref{tab:cifar-accuracy} summarizes the results of our method and various baseline methods. 
    \begin{itemize}
        \item {\bf No Adaptation} attains very poor accuracy, showing that on-device training is necessary. 
        \item {\bf FullTrain} fine-tunes the whole model with BP, and it serves as a strong baseline with the assumption of unlimited memory and computation resources. While {\it FullTrain} achieves excellent accuracy in the FP32 format, its INT8 variant experiences remarkable accuracy drop due to the difficulty of handling a real-quantized model in BP. 
        \item {\bf MCUTrain}~\cite{lin2020mcunet} is a state-of-the-art (SOTA) on-device training method under an extremely low memory budget. This baseline method determines the set of layers/channels to update before an MCU deployment. The
\textit{MCUTrain} method performs better than the IN8 \textit{FullTrain} approach, but still experiences remarkable (around 7-10\%) accuracy drop in certain corruption types (shot noise, impulsive noise, etc) compared with FP32 {\it FullTrain}. 
Ref. \cite{lee2022surgical} showed that fine-tuning the layers where distribution shifts happen can achieve the best performance. Since corruptions lead to distribution shifts in the starting layers, one has to fine-tune these early layers with a significant memory overhead in the BP process.

\item {\bf Our BP-free method} achieves the best accuracy among all on-device training methods in the INT8 format, and it even can match the performance of FP32 \textit{FullTrain}. Since our method train the {\bf full model}, it significantly outperforms the \textit{MCUTrain} method.
    \end{itemize} 

    \item {\bf Hardware Cost (Memory, Computing and Latency).} 
Next, we evaluate the hardware cost of training the MCUNet on an MCU. The peak memory, computation cost and latency of one MCUNet inference are 190 KB, 22.5 MMAcs and 190 ms, respectively. We evaluate single-batch training with no gradient accumulation. The query budget of BP-free training is $Q=50$ for each layer. We perform the same on-device task-adaptive layer selection (Block 3) for both BP-based and BP-free sparse training. The detailed results of full-model training and sparse training are listed in Table \ref{tab:cifar-computation} and Table \ref{tab:cifar-sparse}, respectively. We summarize the key information below. 
\begin{itemize}
    \item {\bf Peak Memory.} The peak memory of our BP-free training remains the minimum possible training memory (trainable parameters + peak inference memory) in both full-model and sparse training. In contrast, full-model BP training easily runs out of memory (3,650 KB) due to the extra memory for saving large activation values in starting layers. 
    \item {\bf Latency \& Training Time.} Our BP-free training has a longer latency and training time, as it needs $Q=50$ forward evaluations per layer to reduce the ZO gradient variance so that our method can converge after the same training iterations of a BP-based method. Reducing $Q$ can reduce the latency per iteration, but will result in more iterations to converge. This trade-off results in a much larger per-step latency and longer end-to-end training time when using the general TF-Lite backend for training.
    \item {\bf Accelerated Performance.} Due to the BP-free nature, our training framework can be easily accelerated by using any existing inference accelerators. For instance, by using the SOTA TinyEngine \cite{lin2020mcunet} for inference, the per-iteration latency of our BP-free training could be greatly reduced, leading to a greatly reduced total training time with BP-based training. While it is also possible to optimize the latency of BP, extra memory as well as extra high-precision computation resources are always needed. By further incorporating the sparse training, we could achieve a comparable end-to-end training time as BP full-model training. 
\end{itemize}
\item {\bf Remark.} BP sparse training could achieve a low training memory~\cite{lin2022device} to fit 256-KB SRAM budget. However, this approach is not fully on-device, as it relies on \textit{cloud} computation to select trainable parameters and perform BP computation graph optimization. Besides, the trainable parameters search space is constrained within ending layers, as training starting layers with BP also necessitates larger memory. Our BP-free training is all-on-edge, enables task-adaptive sparse training, and enables training any layer in the network to achieve the best sparse training performance. The trade-off among memory, flexibility, design/deployment difficulty, and end-to-end training time depends on the actual settings.
\end{itemize}

\begin{table}[htbp]
  \centering
  \caption{Comparison of the hardware cost of full-model training. 
  }
    \begin{tabular}{cccccc}
    \toprule
    \toprule
          &       & \multicolumn{4}{c}{MCUNet			} \\
    \cmidrule{3-6}Backend & Method & Memory (KB) & Compute (MMACs) & Latency (ms) & End-to-end Time (s) \\
    \midrule
    TF-Lite~\cite{david2021tensorflow} & BP    & 3,650  & 76.2  & 13,398 & 6.70E+05 \\
          & BP-free & 668   & 7,857.5 & 428,496 & 2.14E+07 \\
    \midrule
    TinyEngine~\cite{lin2020mcunet} & BP-free & 668   & 7,857.5 & 72,839 & 3.64E+06 \\
    \bottomrule
    \bottomrule
    \end{tabular}%
    
  \label{tab:cifar-computation}%
\end{table}%

\begin{table}[htbp]
  \centering
  \caption{Comparison of the hardware cost of sparse training. The \underline{underlined} are measured results on an STM32F746 MCU.}
    \begin{tabular}{cccccc}
    \toprule
    \toprule
          & 	     & \multicolumn{4}{c}{MCUNet} \\
\cmidrule{3-6}    Method & Test Accu. & Memory (KB) & Compute (MMACs) & Latency (ms) & End-to-end Time (s) \\
    \midrule
    BP~\cite{lin2022device}    & 61.78 & 3,650 & 40.4  & \underline{495}   & 2.48E+04 \\
    BP-free & 62.33 & \underline{209}   & 1,162.5 & \underline{9,939}  & 4.97E+05 \\
    \bottomrule
    \bottomrule
    \end{tabular}%
  \label{tab:cifar-sparse}%
\end{table}%

\subsubsection{Results on the FGVC Dataset}
Table \ref{tab:fgvc-accuracy} summarizes the testing the accuracy of our BP-free method and various BP-based training.
The FGVC dataset mainly perceive output-level distribution shifts, thus training some last layers could match or even surpass full model training \cite{lee2022surgical, lin2022device}. Nevertheless, due to the extremely limited memory budget, the BP-based method can only afford training three middle layers~\cite{lin2022device, zhu2023pockengine} even if exhaustive optimization tricks (e.g., compile-time differentiation, backward graph pruning, operator reordering) applied. This is enough for some easy datasets like Flowers, but insufficient for more challenging datasets including Cars. Our BP-free method can train all layers with no extra memory cost. Therefore, even with a large ZO gradient estimation variance, our method still outperforms BP-based training on most datasets (Cars, CUB, and Pets).


\begin{table}[htbp]
  \centering
  \caption{Validation accuracy comparison on transfer learning to 5 FGVC datasets with our BP-free method and other BP training baselines.  The training settings that could be implemented on an MCU with 256 KB SRAM are \underline{underlined}. Other training settings are out-of-memory and only listed for comparison.}
    \begin{tabular}{cc|cccccc}
    \toprule
    \toprule
          &       & \multicolumn{6}{c}{Accuracy (\%) (MCUNet with 480KB paramemeters)} \\
\cmidrule{3-8}    Method & BP type & Cars  & CUB   & Flowers & Food  & Pets  & Average \\
    \midrule
    Full  & FP32  & 56.7  & 56.2  & 88.8  & \textbf{55.7} & 79.5  & \textbf{67.4} \\
    TinyTrain~\cite{kwon2023tinytrain} & FP32  & 55.2  & 57.8  & \textbf{89.1}  & 52.3  & 80.9  & 67.1 \\
    \underline{MCUTrain}~\cite{lin2022device} & INT8  & 54.6  & 57.3 & 88.1  & 52.1  & 81.5  & 66.7 \\
    \midrule
    \underline{BP-free (Ours)} & /     & \textbf{60.1} & \textbf{58.1}  & 85.0 & 45.5  & \textbf{81.6} & 66.1 \\
    \bottomrule
    \bottomrule
    \end{tabular}%
  \label{tab:fgvc-accuracy}%
\end{table}%



\section{Related Work}

In this section, we review some prior works related to this paper.
\subsection{On-device Training}
On-device inference has been well studied to deploy a pre-trained deep neural network (DNN) for inference on the edge. Driven by the demand to adapt edge-deployed neural network models to new data/new tasks \cite{cai2020tinytl, lin2022device, kwon2023tinytrain} or to unseen distribution shifts at test time \cite{wang2020tent, wang2022continual}, there have been increasing interests in DNN training on edge devices.
However, edge devices usually have tight memory and computation resources, and run without an operating system, making it infeasible to implement standard deep learning training frameworks that rely on automatic differentiation~\cite{baydin2018automatic}. To implement training with BP on edge devices one has to implement gradient computations by hand, which is time-consuming and needs specific optimization.
Moreover, BP-based training requires extra memory to save intermediate results as well as high-precision computation.
The most memory-efficient scheme is to skip all weight updates (e.g., fine-tuning only the last layer \cite{mudrakarta2018k, ren2021tinyol}, bias only \cite{cai2020tinytl}, normalization parameters \cite{frankle2020training}), yet such a scheme leads to considerable accuracy drop compared with full-model training.
As a result, it is almost infeasible to support training on the same device that only supports inference.
\textit{MCUTrain} \cite{lin2022device} enabled training on a microcontroller with merely 256KB SRAM and matched cloud training results on the VisualWakeWords dataset. However, \textit{MCUTrain} has a significant cloud computation overhead, requiring thousands of runs of evolutionary search to find the best weight update scheme that fits the memory budget. \textit{MCUTrain} also needs compilation-time optimization to reduce the peak memory during BP. As a result, this method is not fully on-device training: with a large cloud overhead, it can only manage to train 4 layers (out of 42). \textit{TinyTrain} \cite{kwon2023tinytrain} further enabled on-device parameter selection and showed comparable performance to full-model training on some vision classification tasks. However, to implement the update scheme on edge devices without an operating system (e.g., MCU, FPGA), compilation-time optimization is still needed. As a result, to achieve comparable training performance, the current \textit{TinyTrain} method is still not fully on-device training. In summary, state-of-the-art BP-based on-device training can only afford training some last layers due to the memory constraints.

\subsection{BP-Free Training}
Due to the challenge of implementing BP on edge devices, several BP-free training algorithms have been proposed. These methods have gained more attention in recent years as BP is also considered “biologically implausible”. Zeroth-order (ZO) optimization \cite{nesterov2017random, duchi2015optimal, cai2021zeroth, liu2018zeroth} plays an important role in signal processing and adversarial machine learning where actual gradient information is infeasible (\textit{e.g.}, black-box attack \cite{chen2017zoo, tu2019autozoom, cheng2019improving}). 
Recently, ZO optimization is also applied in neural network training. However, due to the high variance of ZO gradient estimation, previous work focusing on adapting low-dimensional auxiliary modules including input/feature reprogramming \cite{tsai2020transfer, yang2022rep, wang2023soda}, prompt/adapter tuning for emerging large language \cite{guo2023black, yang2024loretta, yang2024adazeta}, vision \cite{oh2023blackvip, niu2024test}, and vision-language models \cite{ouali2023black}, but not the backbone parameters of a pre-trained model.
Recently, \cite{malladi2023fine, zhang2024revisiting} showed that ZO optimization was effective for full-parameter fine-tuning of large language models (LLM) up to 66 Billion parameters. However, this is only applicable when the pre-trained model has a very low intrinsic dimension (200-400) \cite{aghajanyan2020intrinsic, malladi2023kernel}, which is only observed in LLMs. For more general cases, the scalability of ZO training remains a fundamental challenge. Ref. \cite{zhao2023tensor} scaled up ZO training from scratch by leveraging tensor-compressed training for dimension reduction. DeepZero \cite{chen2023deepzero} further scaled up ZO training to train a ResNet-20 with 270K parameters from scratch. However, DeepZero relies on low-variance coordinate-wise gradient estimation which is extremely computation-inefficient thus not suitable for on-device training. ZO optimization also provides a promising solution to training neural networks on emerging computing platforms (e.g., optical neural networks (ONN) \cite{gu2020flops, gu2021efficient, gu2021l2ight, zhao2023real}) where BP is infeasible due to the non-differentiability or limited observability of the analog hardware. Other BP-free training methods include forward-forward algorithm ~\cite{hinton2022forward} for biologically-plausible learning, forward gradient method \cite{baydin2022gradients, fournier2023can, ren2022scaling} based on
forward-mode AD for memory-efficient gradient computation or to avoid stacked BP \cite{cho2024separable, yu2024separable} when higher-order derivatives are needed, log-likelihood method \cite{jiang2023one}, node perturbation \cite{hiratani2022stability, dalm2023effective}, feedback alignment method \cite{nokland2016direct} and input-weight alignment method \cite{chiang2022loss}. Scalability, training stability, and convergence remain the top challenges of all emerging BP-free training frameworks.

\section{Conclusion}
In this paper, we have proposed a quantized BP-free training framework on MCU. With this framework, a quantized inference engine can be easily converted to a training engine. 
Our method leverages quantized ZO optimization to achieve full-model training on a MCU at the similar memory cost of inference. This approach has addressed the long-standing memory challenge that prevents realistic training on a MCU. 
To tackle the slow convergence commonly associated with ZO optimization, we have introduced several innovations, including learning-rate scaling, dimension-reduction techniques such as layer-wise node perturbation, and task-adaptive sparse training. These enhancements have stabilized the training process and improved the convergence.

Our BP-free methods have demonstrated superior training performance over existing BP-based training solutions, primarily due to its full-model training capability under low memory budget and the ability to adapt the layer selection dynamically to specific tasks on-device. This makes the framework highly suitable for vast real applications, where minimal hardware complexity and maximum flexibility in task adaptation are essential, opening the door to broader adoption of on-device training in resource-limited environments.

However, our BP-free training method needs more training iterations than BP-based methods, resulting in longer end-to-end training times. Future research directions include the development of various novel techniques to further improve the convergence. Additionally, investigating the adaptability of BP-free training across various neural network architectures, application domains, and platforms (e.g., integrated photonics, probabilistic circuits) will further expand its applicability, potentially impacting many application areas.

\section*{Acknowledgments}
This work is supported by funding of Intel Strategic Research Sector (SRS) - Emerging Technology.

\bibliographystyle{unsrt}
\bibliography{main}

\newpage
\section*{Appendix}
\appendix

\section{Pseudo-algorithm of Memory-efficient Layer-wise Weight/Node Perturbation}
\begin{algorithm}[h]
  \caption{Memory-efficient Layer-wise Weight/Node Perturbation}
  \label{alg: Me}
\begin{algorithmic}[1]
\Require
    Loss function $\ell(\cdot)$, training dataset $\mathcal{X}$, batch size $N$, total iterations \textit{T}, learning rate schedule ${\eta_t}$,  trainable layers $\{f^{(i)}(\mat{a}^{(i)}, \bm{\theta}^{(i)})\}_{i=0}^L$
\For {$t \leftarrow 0\cdots T-1$}
    \State Mini-batch training data samples $\mat{x}: \{\mat{x}_n\}_{n=1}^N \in \mathcal{X}$
    \State {$\bar{\mat{a}}_0 = \mat{x}$}
    \State $\bm{l} \leftarrow \ell(\bm{\theta}_t;  \mat{x}$) \Comment{Clean Forward}
    \For {$i \leftarrow 0\cdots L-1$}
        \State $\bar{\mat{a}}^{(i+1)}=f^{(i)}(\bar{\mat{a}}^{(i)}, \bar{\bm{\theta}}^{(i)})$
        \If {$d_W < d_a$}   \Comment{Weight Perturbation}
            \State Sample random seed $S$
            \For {$q \leftarrow 0\cdots Q-1$}
                \State $\bm{l}_{q}\leftarrow \ell(\bm{\theta}^{(i)}_t + \mu \bm{\xi}_q;  \bar{\mat{a}}^{(i)})$ \Comment{Generate $\bm{\bar{\xi}}_q$ with S, in-place addition}
            \EndFor

            \State Reset random number generator with seed $S$  
            
            \For {$q \leftarrow 0\cdots Q-1$}
                \State Re-generate $\bm{\bar{\xi}}_n$ with S
                \State $\hat{\nabla}_{\bm{\bar{\theta}}_{i}} \leftarrow \sum_{q} (\bm{l}_{q} - \bm{l}) \bm{\xi}_q$
            \EndFor
        \Else \Comment{Node Perturbation}
            \State Sample random seed $S$
            \For {$q \leftarrow 0\cdots Q-1$}
                \For {$n \leftarrow 0\cdots N-1$}
                    \State $\bar{\mat{a}}^{(i+1)}_{n} \leftarrow \bar{\mat{a}}^{(i+1)}_{n} + \mu \bm{{\xi}}_{q,n}$ \Comment{Generate $\bm{\bar{\xi}}_n$ with S, in-place addition}
                \EndFor
                \State $\bm{l}_{q,n}\leftarrow \ell(\bm{\theta}^t;  \bar{\mat{a}}^{(i+1)}_n)$ \Comment{Partial Forward}
            \EndFor
             
            \State Reset random number generator with seed $S$  
            \For {$q \leftarrow 0\cdots Q-1$}
                \For {$n \leftarrow 0\cdots N-1$}
                    
                    \State Re-generate $\bm{\bar{\xi}}_n$ with S \Comment{Reuse $\bar{\mat{a}}^{(i+1)}$ to store $\hat{\nabla}_{\mat{\bar{a}}^{(i+1)}} \ell$}
                    
                    \State $\hat{\nabla}_{\bm{\bar{\theta}}_{i}} \ell \leftarrow  \hat{\nabla}_{\bm{\bar{\theta}}_{i}} \ell + (\mat{\bar{a}}_{n}^{(i)})^T \bm{{\xi}}_{q,n} (\bm{l}_{q,n}-\bm{l}_n) / \mu$ \Comment{$(\mat{\bar{a}}^{(i)})^T \bm{\bar{\xi}}_n$ can be reduced to INT8 additions}
                \EndFor
            \EndFor
        \EndIf
        
        \State $\bm{\theta}_{t+1}^{(i)} \leftarrow \bm{\theta}_{t}^{(i)} - \eta_t \hat{\nabla}_{\bm{\bar{\theta}}^{(i)}}\ell$ 

        \State $\mat{\bar{a}}^{(i+1)} \leftarrow f^{(i)}(\mat{\bar{a}}^{(i)}, \bm{\theta}_{t}^{(i)})$ \Comment{Recover activation at step $t$}
        
\EndFor
\EndFor
\end{algorithmic}
\end{algorithm}

\section{Quantization-aware Scaling \cite{lin2022device}} \label{appendix:QAS}

The update rule of quantized parameter $\bar{\bm{\theta}}$ with a specific learning rate $\eta_{\bar{\bm{\theta}}}$:
\begin{equation}
    \bar{\bm{\theta}}_{t+1} = \text{clip}\left( \lceil \bar{\bm{\theta}}_t - \eta_{\bar{\bm{\theta}}} \hat{\nabla}_{\bm{\bar{\theta}}} \mathcal{L}(\bar{{\bm{\theta}}}, s_{\bm{\theta}}) \rfloor \right)
\end{equation}

The update rule of full-precision parameter $\bm{\theta}$ with a global learning rate $\eta$:
\begin{equation}
\begin{aligned}
\bm{\theta}_{t+1} & =\bm{\theta}_{t} - \eta \hat{\nabla}_{\bm{\theta}} \mathcal{L} \\
& =s_{\bm{\theta}} \left( \frac{\bm{\theta}_{t}}{s_{\bm{\theta}}} - \eta \frac{\hat{\nabla}_{\bm{\theta}} \mathcal{L} (\bm{\theta})}{s_{\bm{\theta}}} \right) \\
\bar{\bm{\theta}}_{t+1} \approx \bm{\theta}_{t+1} / s_{\bm{\theta}} & = \left( \bar{\bm{\theta}}_t -\frac{\eta}{s_{\bm{\theta}}} \hat{\nabla}_{\bm{\theta}} \mathcal{L} (\bm{\theta})\right) \\
\bar{\bm{\theta}}_{t+1} \approx \bm{\theta}_{t+1} / s_{\bm{\theta}} & = \left( \bar{\bm{\theta}}_t -\frac{\eta}{s_{\bm{\theta}}^2} \hat{\nabla}_{\bm{\bar{\theta}}} \mathcal{L}(\bar{{\bm{\theta}}}, s_{\bm{\theta}})\right) 
\end{aligned}
\end{equation}

Given $\hat{\nabla}_{\bm{\theta}} \mathcal{L} (\bm{\theta})$, update $\bm{\theta}$ with a global learning rate $\eta$ is equivalent to update $\bar{\bm{\theta}}$ with a learning rate ${\eta} / {s_{\bm{\theta}}}$, neglecting the rounding error of $\text{clip}$.

Scaling the learning rate is equivalent to scale the gradient $\hat{\nabla}_{\bm{\bar{\theta}}} \mathcal{L}(\bar{{\bm{\theta}}}, s_{\bm{\theta}})$ by $s_{\bm{\theta}}^{-2}$ at each step so as to unify the hyperparameter $\eta$ as a global learning rate for all layers. The update of real-quantized parameters with different $s_{\bm{\theta}}$ follows the update of full-precision parameters.

\section{Extented Details of MCU Implementation}

The following steps outline the implementation of XORShift to generate random numbers following the Rademacher distribution:

\begin{enumerate}
    \item \textbf{Initialize the seed:} Set the seed for the XORShift generator, which will serve as the initial state.
    \item \textbf{Update the state:} Apply the XORShift recurrence relations to the current state to generate a new pseudo-random number.
    \item \textbf{Generate Rademacher-distributed values:} The generated pseudo-random number is transformed into a Rademacher-distributed value by examining the least significant bit (LSB) of the number. If the LSB is $1$, return $-1$; otherwise, return $+1$.
\end{enumerate}

The C++ code snippet below demonstrates how to implement XORShift to generate Rademacher-distributed random numbers:

\begin{verbatim}
unsigned int xor_seed; // XORShift seed

// Set the XORShift seed
void set_xor_seed(unsigned int s) {
    xor_seed = s;
}

// XORShift pseudo-random number generator
unsigned int xor_rand() {
    xor_seed ^= xor_seed << 13;
    xor_seed ^= xor_seed >> 17;
    xor_seed ^= xor_seed << 5;
    return xor_seed;
}

// Generate a Rademacher-distributed value
int rademacher() {
    unsigned int rand_num = xor_rand();
    return (rand_num & 1) ? -1 : 1;
}
\end{verbatim}

In this implementation, the function \texttt{xor\_rand()} generates a pseudo-random number using the XORShift algorithm, and the function \texttt{rademacher()} converts this number into a Rademacher-distributed value by checking the least significant bit (LSB).

\end{document}